\documentclass[pdflatex,sn-basic]{sn-jnl}
\jyear{2022}%

\renewenvironment{table}[1][]%
{\tableorg[#1]%
\tablebodyfont%
\renewcommand\footnotetext[2][]{{\removelastskip\vskip3pt%
\let\tablebodyfont\tablefootnotefont%
\hskip0pt\if!##1!\else{\smash{$^{##1}$}}\fi##2\par}}%
}{\endtableorg}

\theoremstyle{thmstyletwo}%

\theoremstyle{thmstylethree}%
\newtheorem{definition}{Definition}%

\usepackage{url}

\usepackage{graphicx}
\usepackage{float}
\usepackage{amssymb}
\usepackage{amsmath,amssymb}

\usepackage{xspace}
\usepackage{xparse} 

\usepackage{algorithm}
\usepackage{algpseudocode}
\usepackage{ulem}

\usepackage{bm}

\DeclareMathOperator{\theargmin}{argmin}

\DeclareMathOperator{\theSBM}{\mathcal{SBM}}
\DeclareMathOperator{\SBM}{SBM}
\DeclareRobustCommand{\sbm}[1]{{\SBM\mspace{-1mu} ( \mspace{-2mu} #1 \mspace{-1mu} )}}

\newcommand{\eqdef}{\stackrel{\text{\tiny def}}{=}}

\newcommand{\bA}{\bm{A}}
\newcommand{\bAp}{\bm{A}^\prime}
\newcommand{\bAk}{\bm{A}^{(k)}}
\newcommand{\bD}{\bm{D}}
\newcommand{\bI}{\bm{I}}
\newcommand{\bL}{\bm{L}}

\newcommand{\blb}{\bm{\lambda}}
\newcommand{\bFi}{\bm{\Phi}}
\newcommand{\btau}{\bm{\tau}}

\newcommand{\bDlt}{\bm{\Delta}}
\newcommand{\avgDel}{{\overline{\bm{\Delta}_{\blb}}}}
\newcommand{\avgDelP}{{\overline{\bm{\Delta}^p_{\blb}}}}

\newcommand{\bB}{\bm{B}}

\newcommand{\bP}{\bm{P}}

\newcommand{\cG}{\mathcal{G}}

\newcommand{\cS}{\mathcal{S}}

\newcommand{\gnP}{\mathcal{G}\mspace{-1mu}\big(\mspace{-2mu}n,\bP\mspace{-1mu}\big)}
\newcommand{\Gk}{G^{(k)}}

\DeclareMathOperator{\pr}{\mathbb{P}}

\DeclareRobustCommand{\argmin}[1]{\underset{#1}{\theargmin}\mspace{4mu}}

\DeclareRobustCommand{\fm}[1]{\bm{\mu} \mspace{-2mu} \big [\mspace{-2mu} #1 \mspace{-2mu}\big]}                 

\DeclareRobustCommand{\sE}[1]{\widehat{\mathbb{E}}\mspace{-2mu}\left[#1 \right]} 

\DeclareRobustCommand{\prob}[1]{\mspace{2mu}\mathbb{P}\mspace{-2mu}\left(\mspace{-2mu} #1 \mspace{-2mu}\right)}

\NewDocumentCommand
\nfm {o}   
     {
       \IfNoValueTF {#1} {\widetilde{\bm{\mu}}_N}{\widetilde{\bm{\mu}}_N \mspace{-1mu} [\mspace{-1mu}#1 \mspace{-1mu}]}
     }
\NewDocumentCommand
\sfm {o}   
     {
       \IfNoValueTF {#1} {\widehat{\bm{\mu}}_N}{\widehat{\bm{\mu}}_N \mspace{-1mu} [\mspace{-1mu}#1 \mspace{-1mu}]}
     }
     
\NewDocumentCommand
\ofm {o}   
     {
       \IfNoValueTF {#1} {{\bm{\mu}}^{\ast}_N}{{\bm{\mu}}^{\ast}_N \mspace{-1mu} [\mspace{-1mu}#1 \mspace{-1mu}]}
     }
     
\newcommand{\lFM}{{\blb(\widehat{\bm{\mu}}_N)}}
\newcommand{\olF}{{\blb(\bm{\mu}^{\ast}_N)}}

\newcommand{\EA}{\sfm[\bA]}       
\newcommand{\oEA}{\ofm[\bA]}       

\newcommand{\EG}{\sfm[G]}         
\newcommand{\oEG}{\ofm[G]}        

\DeclareRobustCommand{\dA}[1]{d_{\bA} \big(#1\big)} 
\DeclareRobustCommand{\dL}[1]{d_{\bL} \big(#1\big)} 
\DeclareRobustCommand{\dH}[1]{d_{H} \big(#1\big)} 

\newcommand{\ER}{Erd\H{o}s-R\'enyi\xspace}
\newcommand{\fr}{Fr\'echet\xspace}

\newcommand{\citeg}[1]{(e.g., \citep{#1} and references therein)}

\begin{document}
\title[Estimation of Sample \fr Mean]{Estimation of the Sample \fr Mean:\\ A Convolutional Neural Network Approach}
\author[1]{\fnm{Adam} \sur{Sanchez}}\email{adam.sanchez@colorado.edu}
\author*[1]{\fnm{Fran\c{c}ois G.} \sur{Meyer}}\email{fmeyer@colorado.edu}
\affil[1]{\orgdiv{Department of Applied Mathematics}, \orgname{University of Colorado
    Boulder}, \orgaddress{\street{525 UCB}, \city{Boulder}, \postcode{80309-0526}, \state{CO}, \country{USA}}}

\abstract{ This work aims to address this rising demand for novel tools in statistical and
  machine learning for ``graph-valued random variables'' by proposing a fast algorithm to
  compute the sample \fr mean, which replaces the concept of sample mean for graphs (or
  networks).  We use convolutional neural networks to learn the morphology of the graphs
  in a set of graphs. Our experiments on several ensembles of random graphs demonstrate
  that our method can reliably recover the sample \fr mean.}

\maketitle

\section{Introduction}
The purpose of this paper is to present a fast method to compute the sample \fr mean graph
using convolutional neural networks. The \fr mean (or median) graph, which extends the
notion of mean to probability measures defined on metric spaces \citep{frechet47}, has
become a standard tool for the analysis of graph-valued data
\citeg{dubey20,ferrer10,ginestet17,jain16b,josephs21,kolaczyk20,lunagomez20}. The vital
role played by the \fr mean as a location parameter \citep{jain12,jain16a,kolaczyk20}, is
exemplified in the works of \citep{banks98,lunagomez20}, who have created novel families
of random graphs by generating random perturbations around a given \fr mean graph.
\subsection{Related work}
The \fr mean graph has been studied in the context where the distance that is used to compare graphs is the edit
distance \citeg{bardaji10a,bardaji10b,ginestet12,jain09,jiang01}, the Frobenius distance
\citeg{boria20,jain08,kolouri21}, and the Gromov-Wasserstein distance
\citeg{barbe20,chowdhury19,gu15,heinemann22,metivier19,vayer20}. In this paper, we propose to compare graphs using the
eigenvalues of their respective adjacency matrices.

The \fr mean graph is the solution to a minimization problem that is intractable for most distances
\citeg{anderes16,bardaji10a,heinemann22,jain09,patterson21}. For this reason several alternatives have been proposed to
solve the minimization problem (\ref{sample-frechet-mean}). A popular approach consists in embedding the graphs in a
Euclidean space, wherein one can trivially find the mean of the set \citeg{ferrer10,kolouri21,simou20}.  Several
researchers have proposed recently to learn the embedding from massive datasets of existing networks. Such algorithms
learn an embedding \citeg{preuer18} from a set of graphs into Euclidean space, and then compute a notion of similarity
between the embedded graphs. Several authors have proposed to use convolutional neural networks to compute the
intractable Wasserstein distance \citeg{brogat22,solomon15}.

When the graphs $\left\{ \Gk \right\}_{1\leq k \leq N}$, in the sample are not defined on the same vertex set, then one
needs to solve a graph isomorphism problem to jointly align the graphs \citeg{ferrer10}. In this paper, we assume that
the graphs are defined on the same vertex set, and we avoid the added combinatorial complexity associated with the graph
alignment problem.
\subsection{The \fr mean graph}
Let $\cG$ be the set of all simple labeled graphs with vertex set $\left\{1, \ldots ,n\right\}$, and let $\cS$ be the set of
$n \times n$ adjacency matrices of graphs in $\cG$,
\begin{equation}
  \cS = \left \{
  \bA \in \{0,1\}^{n \times n}; \text{where} \; a_{ij} = a_{ji},\text{and}  \; a_{i,i} = 0; \; 1 \leq i < j \leq n
  \right\}.
  \label{adjacency_matrices}
\end{equation}
We equip $\cG$ with a metric $d$ to measure the distance between two graphs. We characterize the ``average'' of a sample
of graphs $\left\{G^{(1)}, \ldots, G^{(N)}\right\}$, which are defined on the same vertex set $\left\{1, \ldots
,n\right\}$, with the sample \fr mean graphs \citep{frechet47}.
\begin{definition}
  The sample \fr mean graphs are solutions to 
  \begin{equation}
    \EG =  \argmin{G\in \cG} \frac{1}{N}\sum_{k=1}^N d^2(G,\Gk).
    \label{sample-frechet-mean}
  \end{equation}
\end{definition}
Solutions to the minimization problem (\ref{sample-frechet-mean}) always exist, but the minimizers need not be
unique. All our results are stated in terms of any of the elements in the set of minimizers of
(\ref{sample-frechet-mean}). To simplify the exposition, we refer to the sample \fr mean graph as any representative
element of the class of solutions to (\ref{sample-frechet-mean}), and with a slight abuse of notation (which is commonly
used), we denote by $\EG$ this representative element.
\subsection{Our main contributions}
In this paper, we propose a novel method for calculating the sample \fr mean $\EG$.  Our line of attack relies on the
following key ideas: (1) stochastic block models with community of various sizes provide universal approximants to
graphs \citep{bickel09,ferguson22a,olhede14}, (2) computing the sample \fr mean of stochastic block models (sampled from
the same probability measure) can be performed using simple averaging and nonlinear thresholding \citep{meyer22}. We
propose therefore to combine these two principles within a convolutional network, which learns the combined optimal
(stochastic block model) approximation of the graphs in a sample, together with the averaging and nonlinear thresholding
that yields the sample \fr mean.

We show in many experiments with various ensembles of random graphs, that the estimate of the \fr mean computed with our
method is very close to the true \fr mean (computed using a brute force approach) when the distance used to compare
graphs is the adjacency spectral  pseudo-distance. This property is a consequence of the following fact: the stochastic
block model that provides a nonlinear approximation to all the graphs in the sample is able to capture the large scale
connectivity features of the \fr mean graph. Conversely, we also show that the local connectivity structure (as
quantified by the degree distribution) of the true \fr mean is very similar to the corresponding connectivity structure
of the graph computed with our method.

This experimental result is significant because the convolutional neural network does not require the computational
costly computation of the eigenvalues of the graphs in the sample.
\subsection{Organization of the paper}
In the next section, we introduce the main concepts and associated notations. In section \ref{overview}, we provide an
overview of our approach. The architecture of the convolutional neural network is described in section \ref{CNN}.  The
random graph ensembles that are used to train the convolutional neural network are presented in section
\ref{train-section}. In the same section, we describe the training procedure. Results of experiments conducted on
several ensembles of random graphs are presented in section \ref{experiments-section}.  All relevant code, including the
pre-trained models, is available at \citep{frechet-adam-sanchez22}.

\section{Preliminary and Notations}
For a graph $G \in \cG$, we denote by $\bA$ its adjacency matrix. We define the degree of a vertex, $i \in V$ as $d_i =
\sum_{i \sim j} 1$. The degree matrix, $\bD$, of graph $G$ is the diagonal matrix defined by
\begin{equation}
  \bD = \text{diag}(d_1,\ldots,d_n)
\end{equation}
and the combinatorial Laplacian matrix, $\bL$, is defined by
\begin{equation}
  \bL = \bD - \bA, \label{laplace}
\end{equation}
where $\bI$ is the $n \times n$ identity matrix.
\begin{definition}
  \label{spectrum}
  We denote by $\blb (\bA)= \begin{bmatrix} \lambda_1(\bA)& \cdots& \lambda_n(\bA) \end{bmatrix}$, the vector of sorted
  eigenvalues of the adjacency matrix $\bA$, with the convention that $\lambda_1 (\bA) \ge \ldots \ge
  \lambda_n(\bA)$.

  Similarly, we denote by $\blb (\bL)$ the vector of sorted eigenvalues {\bf in ascending order} of the
  normalized Laplacian matrix $\bL$, with the convention that $\lambda_1 (\bL) \le \ldots \le \lambda_n(\bL)$.\\
\end{definition}

\subsection{Distances between graphs}
In this work, we propose an algorithm to construct the barycenter of a sample of graphs. The notion of barycenter, or \fr
mean, requires that we define a distance on the set of graphs. The distance quantifies our notion of similarity between
graphs.  To simplify the exposition, we only consider graphs defined on the same vertex set. In practice, one often
needs to compare graphs of different sizes. Because we use spectral distances, we can easily extend theoretically and
numerically these distances to graphs of different sizes, without having to solve the graph isomorphism problem
\citep{mckay2014}.

We recognize that the computation of the sample \fr mean requires that the distance $d$ being used in
(\ref{sample-frechet-mean}) be adapted to the distinctive characteristics (e.g., degree distribution, presence of
communities, hubs, connectivity structure, etc.) of the sample of graphs.  In the absence of a universal distance that
would be optimal for all graphs \citep{Donnat2018,wills20c}, we propose instead to adapt the distance to the
characteristic features of the graph ensembles from which the sample is drawn \citep{Donnat2018,wills20c}. 

The authors in \citep{wills20c} provide a comprehensive analysis of the properties of many graph distances in the context
of graph valued machine learning algorithms.  They show that the adjacency spectral pseudo-distance
(\ref{adjacency-spectral}) defined as follows,
\begin{definition}
  Let $G,G^\prime \in \cG$ be two graphs with adjacency matrix $\bA$ and $\bA^\prime$ respectively.  We define the
  adjacency spectral pseudometric as the $\ell_2$ norm between the vectors of eigenvalues $\blb (\bA)$ and $\blb
  (\bA^\prime)$ of $\bA$ and $\bA^\prime$ respectively,
  \begin{align} 
    \dA{G,G^\prime} = || \blb (\bA) - \blb (\bA^\prime)||_2. \label{adjacency-spectral}
  \end{align}
\end{definition}
provides the largest decision boundaries between a class of stochastic block model graphs and
a class of \ER random graphs. We adopt the adjacency spectral pseudo-distance to quantify the variations within the
ensemble of stochastic block models.

Conversely, the combinatorial Laplacian spectral pseudo-metric, defined by
\begin{definition}
  Let $G,G^\prime \in \cG$ with combinatorial Laplacian matrix $\bL$ and $\bL^\prime$ respectively.  We define the
  combinatorial Laplacian spectral pseudometric as the $\ell_2$ norm between the vectors of eigenvalues $\blb (\bL)$ and
  $\blb (\bL^\prime)$ of $\bL$ and $\bL^\prime$ respectively,
  \begin{align} 
    \dL{G,G^\prime} = || \blb (\bL) - \blb (\bL^\prime)||_2. \label{distance_laplace}
  \end{align}
\end{definition}
was shown in \citep{wills20c} to provide the largest margin when comparing a class of preferential attachment graphs
\citep{Barabasi1999} and a class of random graphs. For this reason, we use the Laplacian spectral pseudo-distance to
quantify the variations within the ensemble of preferential attachment graphs.\\

We note that $d_{\bA}$ and $d_{\bL}$ are only a
pseudo-distance: they satisfy the symmetry and triangle inequality axioms, but not the identity axiom. Instead, the
pseudo-distances satisfy the reflexivity axiom, $\forall G \in \cG$, $d_{\bA}s(G,G) = d_{\bL} =0$. However, an advantage
of these pseudo-distances is that they do not require node correspondence between $G$ and $G^\prime$.

Finally, when comparing inhomogeneous \ER random graphs, we use the Hamming distance, $d_H$, defined as follows.
\begin{definition}
The Hamming distance between $G$ and  $G^\prime$ with adjacency matrix $\bA$ and $\bAp$ respectively, is given by
  \begin{equation}
    \dH{G,G^\prime} = \sum_{1 \leq i< j \leq n} \lvert a_{ij} - a^\prime_{ij}\rvert.
  \end{equation}
    \label{hamming}
\end{definition}
The advantage of the Hamming distance is that the population and sample \fr means can be computed in closed form \citep{meyer22}.
\subsection{The sample \fr mean, our estimate of the \fr mean, and the sample mean adjacency matrix}
Let $\left\{ \Gk \right\}_{1\leq k \leq N}$, be a set of $N$ graphs. We recall and introduce some notations associated
with the computation of the sample \fr mean graph.
In the following we need to compute (using a brute force approach) the true \fr mean, $\EG$, solution to the
discrete optimization problem (\ref{sample-frechet-mean}).

\begin{definition}
  When optimizing the convolutional neural network, in section \ref{CNN}, we call the adjacency matrix of the true \fr
  mean graph, $\EA$, the {\em target} adjacency matrix.\\

  In contrast, we denote by $\oEG$ the estimate of the sample \fr mean graph computed using the method described in this
  paper, and outlined in the next section. The corresponding adjacency matrix is denoted by $\oEA$.\\

  We denote by $\sE{\bA}$ the sample mean adjacency matrix
  \begin{equation}
    \sE{\bA} = \frac{1}{N} \sum_{k=1}^N \bAk.
  \end{equation}
  The sample mean matrix is the {\em input} to the convolutional neural network.

  Finally, we define the the naive sample \fr mean, $\nfm$, to be the thresholded sample mean adjacency matrix,
  \begin{equation}
    [\nfm]_{ij} =
    \begin{cases}
      1 & \text {if} \quad \sE{\bA}_{ij} > 1/2,\\
      0 & \text{otherwise.}
    \end{cases}
  \end{equation}
\end{definition}
Our goal is to demonstrate that $\oEA$ is close to the true sample \fr mean, $\EA$. In order to provide a reasonable
upper bound on the approximation error between the different models and the true sample \fr mean, we define the concept
of ``naive sample \fr mean''. The naive sample mean \fr mean, $\nfm$, only provides a poor man's estimate of sample \fr mean.
Interestingly, the naive sample \fr mean $\nfm$ coincides with the true sample \fr mean $\sfm$ when the graphs in the
ensemble have a reasonably simple geometry \citep{meyer22b}, for instance when all the graphs in the
sample are generated from an inhomogeneous \ER random graph model. 
\section{Overview of our method
  \label{overview}}
\subsection{Computing an approximation to the sample \fr mean}
The optimization problem (\ref{sample-frechet-mean}) is non convex, and in general the computation of the \fr mean is
NP-complete. In this work, we propose a fast method to compute an approximation to the sample \fr mean of a set of $N$ graphs
$\left\{ \Gk \right\}_{1\leq k \leq N}$. Our line of attack relies on the idea of replacing the intractable problem
(\ref{sample-frechet-mean}) with the estimation of a parametric graph model that approximates the sample \fr mean,
$\EA$, arbitrarily well.

The ensemble of stochastic block models, $\theSBM$, with community of various sizes, is a family of parametric model that
provides universal approximants to graphs \citep{ferguson22a,olhede14}. This approximation property of the stochastic
block model holds for various norms: the $\ell^2$ difference between the corresponding spectra
\citep{ferguson22a,ferguson22b}, or the $L^2$ difference between the corresponding graphons
\citep{olhede14,wolfe13}. Different algorithms have been proposed to compute the optimal stochastic block model
approximation to a graph \citep{ferguson22a,olhede14,wolfe13}.

We propose therefore to estimate the SBM graph, $\sbm{\bP}$, with edge probability matrix $\bP$, whose population \fr
mean $\fm{\bP}$, is as close as possible to the sample \fr mean $\EA$. We replace (\ref{sample-frechet-mean}) with
the following optimization program,
\begin{equation}
  \min_{\sbm{\bP} \in \theSBM} \dA{\EA,\fm{\bP}},
  \label{program2}
\end{equation}
where $d_{\bA}$ is the adjacency spectral pseudometric (\ref{adjacency-spectral}), and $\theSBM$ is the set of adjacency
matrices of the stochastic block models. We use the adjacency spectral pseudometric, because we know from
\citep{ferguson22a,ferguson22b} that we can make this distance in (\ref{program2}) asymptotically small in the limit of
large graph sizes.

Each stochastic block model $\sbm{\bP} \in \theSBM$ is a random graph of size
$n$ with edge probability matrix $\bP$ \citep{abbe18}. The set $\theSBM$ is therefore parameterized by the set of $n
\times n$ edge probability matrices $\bP$. 

To solve (\ref{program2}), we first replace the population \fr mean $\fm{\bP}$ of $\sbm{\bP}$ with the sample \fr mean, $\sfm[\bB]$,
computed from a sample of adjacency matrices $\bB^{(k)}$ sampled from the stochastic block models $\sbm{\bP}$.

This approximation is justified by the fact that the sample \fr mean, $\sfm[\bB]$, converges in probability to the
corresponding population \fr mean, $\fm{\bP}$, in the limit of large graph sizes $(n \rightarrow \infty)$,
\citep{meyer22}. The problem (\ref{program2}) is therefore replaced by
\begin{equation}
  \min_{\sbm{\bP} \in \theSBM } \dA{\EA,\sfm[\bB]},\quad \text{where} \quad \bB \sim \sbm{\bP}.
  \label{program3}
\end{equation}
Finally, we take advantage of the following property: the sample \fr mean (computed using the Hamming distance) of a
sample of stochastic block models $\bB \sim \sbm{\bP}$ is obtained by thresholding the sample mean adjacency matrix
$\sE{\bB}$ \citep{meyer22}. In other words,
\begin{equation}
\big[\sfm[\bB]\big]_{ij} =  \big[\btau(\sE{\bB}) \big]_{ij} =
  \begin{cases}
    1 & \text {if} \quad \sE{\bB}_{ij} > 1/2,\\
    0 & \text{otherwise.}
  \end{cases}
  \label{threshold}
\end{equation}
We note that this simple computation of the sample \fr mean does not hold for the adjacency spectral pseudo-distance,
$d_{\bA}$. Notwithstanding our choice of distance, and inspired by the nonlinear rule (\ref{threshold}), we propose
instead to replace $\sE{\bB}$ with $\sE{\bA}$, the sample mean adjacency matrix of the sample $\big\{\bAk\big\},
k=1,\ldots,N$, and to replace the nonlinear thresholding matrix function $\btau$ in (\ref{threshold}) with a more
general nonlinear matrix function $\bFi$. We therefore consider the following optimization program,
\begin{equation}
  \min_{\bFi} d_H \big (\EA,\bFi(\sE{\bA}) \big),
  \label{program5}
\end{equation}
where $\bFi$ is a nonlinear matrix function that operates on the entries of the matrix $\sE{\bA}$. The matrix function
$\bFi$ needs to account for the thresholding $\btau$ introduced in (\ref{threshold}), but may also account for the
approximation errors introduced by replacing the optimization program (\ref{sample-frechet-mean}) with the program
(\ref{program5}).

In the next section, we explain how we can learn the matrix function $\bFi$, which maps the sample mean adjacency matrix
to the sample \fr mean, using a convolutional neural network.

\subsection{Learning the nonlinear matrix function ${\Phi}$}
We propose to use a convolutional network (described in detail in section~\ref{CNN}) to
learn the combined optimal approximation of the graphs in a sample along with the
averaging and nonlinear thresholding that yields the \fr mean. The core engine of the
nonlinear matrix function $\bFi$ is a clustering (segmentation) algorithm that partitions
the sample mean adjacency matrix into modules or communities. Communities are formed by
regions wherein the nodes are densely connected; whereas the other regions correspond to
``non-edge'' entries of the adjacency matrix. The sample mean average adjacency matrix can
be interpreted as a weighted graph; the outcome of the partitioning is an unweighted
graph.

Our approach is similar to the sorting and smoothing
approach of the authors in \citep{chan14} who use Total Variation (TV) denoising to
segment the image formed by the adjacency matrix (re-ordered according to the degree of
each node). The authors in \citep{wei18,wei21} also use image segmentation to estimate the
graphon that is used to generate an observed adjacency matrix. A clustering algorithm is
also used by the authors in \citep{cai15} to estimate a step graphon.

\begin{figure}[H]
  \centerline{
    \includegraphics[width=.5\textwidth]{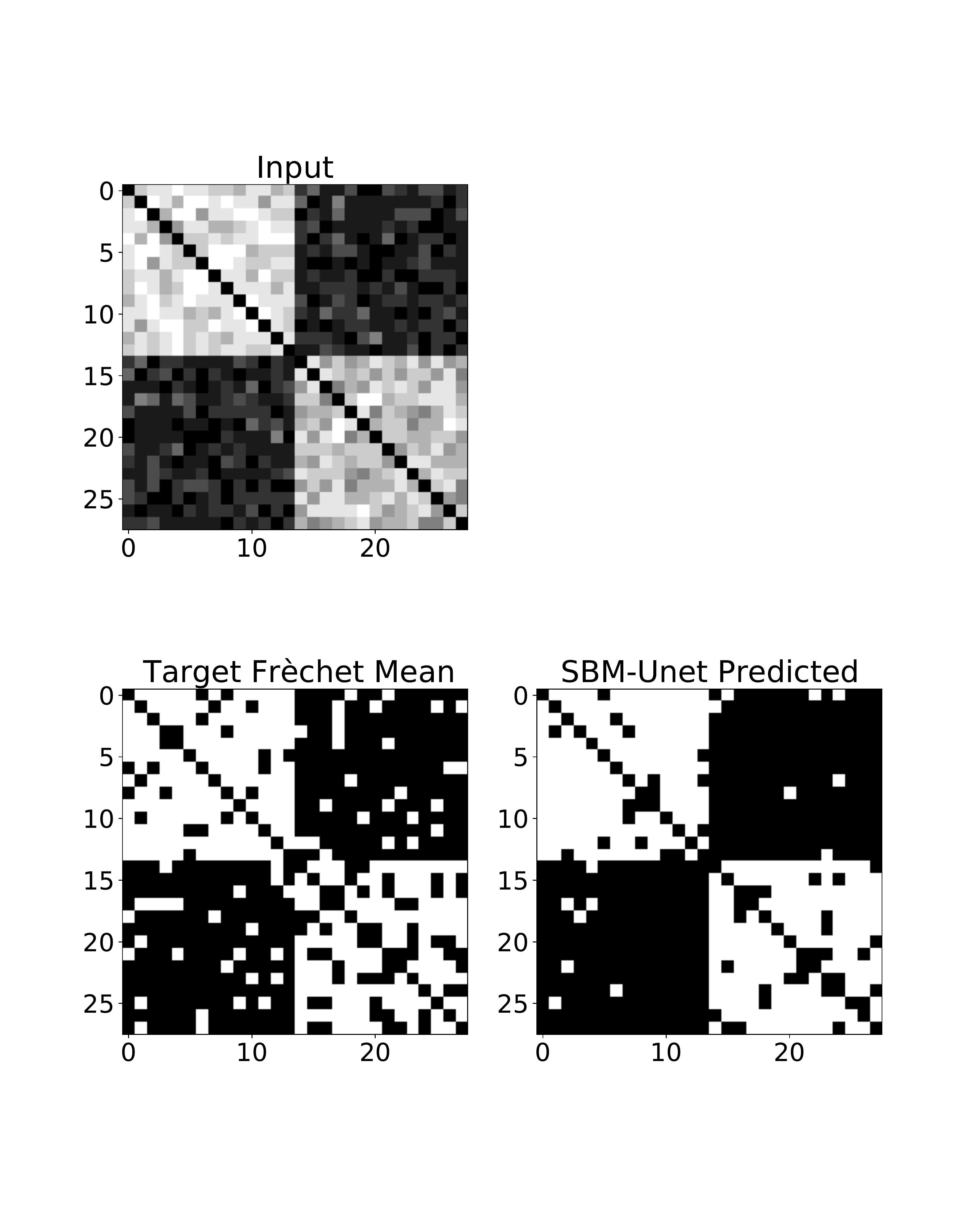}
  }
  \caption{\textbf{Top}: the sample mean adjacency matrix, $\sE{\bA}$, which is the input to the CNN. \textbf{Bottom}:
    Left: a gray scale image of the true sample \fr mean graph's adjacency matrix, $\EA$. Right: $\oEG$, the estimate
    of the \fr mean using our approach.
    \label{fig1}}
\end{figure}

Fig.~\ref{fig1} provides a visual example of the input sample mean adjacency matrix, $\sE{\bA}$ (top-left), the true
sample \fr mean graph's adjacency matrix, $\EA$ (bottom-left), and the estimate of the \fr mean using our approach,
$\oEG$ (bottom-right).

\section{Convolutional Neural Network Framework
  \label{CNN}}
The input to the convolutional neural network is the sample mean adjacency matrix, $\sE{\bA}$, and the output is
$\oEG$. During the training phase, the
\noindent neural network is provided with the true sample \fr mean, $\EA$. The true \fr
mean is computed using a brute force search over the space of undirected unweighted graphs, $\cG$.

When computing $\EA$ during the training or evaluation of the algorithm on a specific graph ensemble, we use the metric
that is best adapted to this ensemble: the Hamming distance $d_H$ for inhomogeneous \ER random graphs, the adjacency
spectral pseudo-distance $d_{\bA}$ for stochastic block-models, and the Laplacian spectral pseudo-distance $d_{\bL}$ for
preferential attachment graphs.

The parameters of the network are then optimized in order to learn the nonlinear map $\bFi$ that computes $\EA$ when
presented with the input $\sE{\bA}$

The design of the architecture of the convolutional neural network relies on the simple idea that $\bFi$, defined in the
program (\ref{program5}), should be able to perform an entry-wise thresholding of the sample mean adjacency matrix
$\sE{\bA}$, defined in (\ref{threshold}), when the input graphs are realizations of inhomogeneous \ER random graphs
(withe the same edge probability matrix $\bP$). We propose to model this nonlinear transformation with a segmentation
algorithm that partitions the sample mean adjacency matrix into modules of densely connected vertices. 
\subsection{U-Nets}
Given the important role segmentation plays in many fields, it is not surprising that
there has been many works dedicated to its advancement. Undoubtedly, the U-Net algorithm
\citep{ronneberger15} has been proven as a key staple in the field. U-Net is a CNN
architecture composed of encoder and decoder structures each with
convolutional/down-sampling and deconvolutional/up-sampling stages. Each of the
convolutional stages are comprised of two $3\times3$ unpadded convolutions followed by a
rectified linear unit (ReLU) and a $2 \times 2$ max pooling operation. At each
down-sampling stage, the number of feature channels are doubled. At each of the
up-sampling stages, an up-sampling of the feature map followed by a $2\time2$ convolution
occurs, halving the number of feature channels. A concatenation with the corresponding
deconvolutional stage is done, followed by two $3\times3$ convolutions and a ReLU. A final
$1 \times 1$ convolution is done in order to achieve the desired output size.

\begin{figure}[H]
  \centering
  \includegraphics[width=\textwidth]{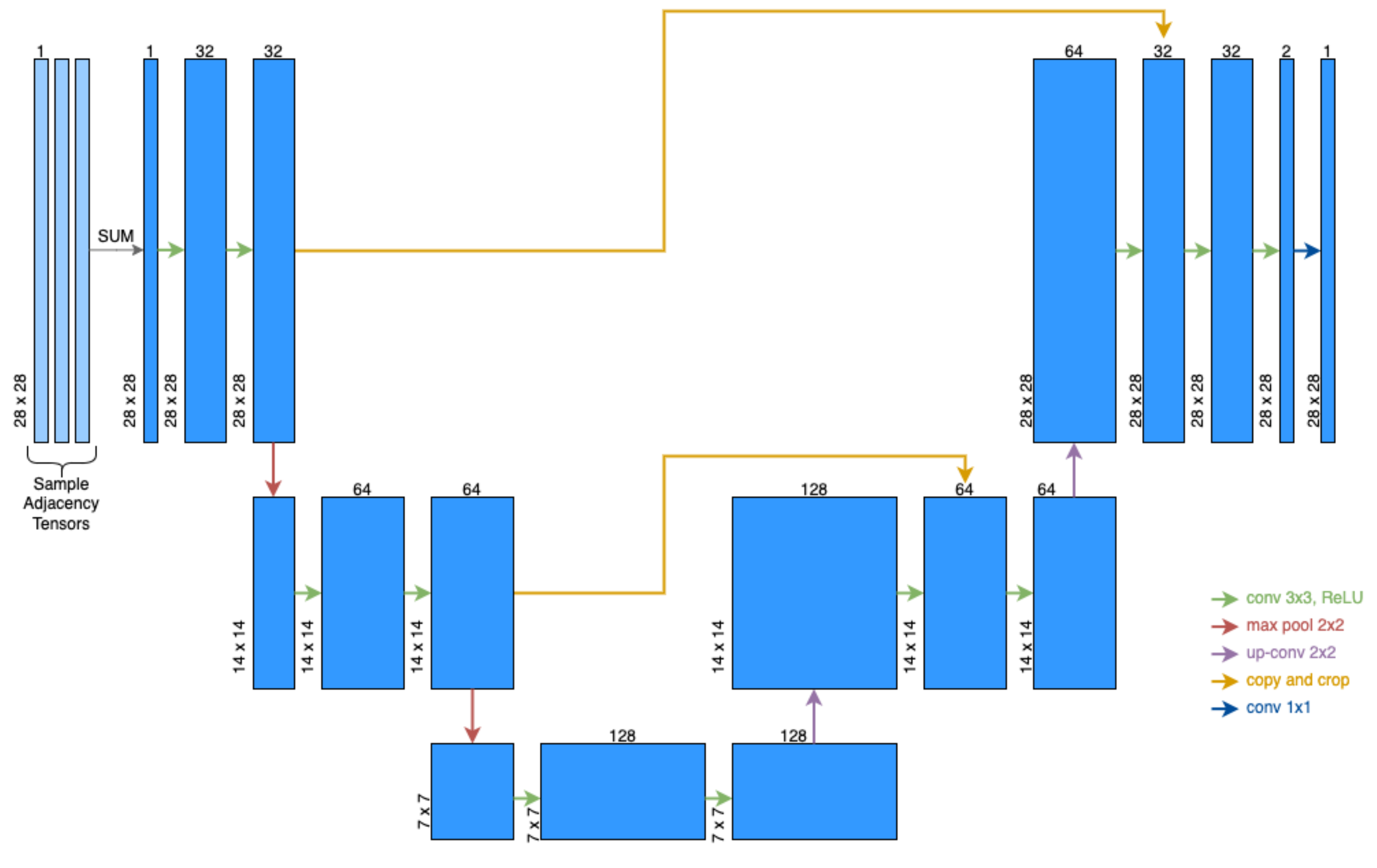}
  \caption{Our model architecture corresponding to a $28 \times 28 \times 1$ input tensor. Each blue box corresponds to
    a multi-channel feature map. The x-y dimensions are given in the lower left of each box. The number of features is
    given by the number above each box. The arrows represent the different operations}
  \label{fig2}
\end{figure}

Our model follows an architecture very similar to U-Net and is depicted in Fig. ~\ref{fig2}. We have a set of 2
down-sampling stages coupled with 2 up-sampling stages, each of which are comprised of $3 \times 3$ convolutions and ReLU
activation. Again, borrowing from U-Net, we add cross concatenations from the down-sampling stages to their
corresponding up-sampling stage. We then perform a final $1\times1$ convolution with a sigmoid activation to achieve our
desired output shape.
\noindent  
\section{Data Sets and Training
  \label{train-section}}
As mentioned above, we restrict ourselves to undirected and unweighted graphs with $n=28$ nodes. We restrict ourselves
to graphs of such small size because our interest is to demonstrate the feasibility of our approach, and therefore we
need to compute the true sample \fr mean graph, solution of (\ref{sample-frechet-mean}). This optimization problem is
solved using an exhaustive search, which is very demanding and unrealistic for large graphs. The computational cost
scales with the sample size, and we therefore restrict ourselves to a sample size of $N = 10$.

We plan to extend our implementation to larger graphs and larger sample size in the future. We note that others have
also evaluated graph-valued machine learning algorithms on graphs of small size (e.g., size of the benchmark graphs for
the comparison and evaluation of graph-valued algorithms \citep{morris20}).

Our code, including the pre-trained models, is available \citep{frechet-adam-sanchez22}.

We consider three distinct ensemble of random graphs in order to generate both our training and testing data:
inhomogeneous \ER random graphs, the stochastic block model, and the Barabasi-Albert preferential attachment model.  We
consider existing ensembles of random graphs as prototypical examples of certain graph \emph{structures}, which are the
building blocks of existing real world networks. Our investigation is concerned with the relationship between the
families of graph ensembles and the ability of one ensemble to provide the best training dataset for the other
ensembles. We expect that the differences between the structural features characteristic of these ensembles 
affect the outcome of the training and the subsequent performance of the algorithm.

For each ensemble of random graphs, we generate a training set of $m=3,600$ graphs from that ensemble. The training
sample is then divided into batches of size $N=10$. For each batch, we compute the exact sample \fr mean graph, $\EA$,
solution to (\ref{sample-frechet-mean}), and the sample mean adjacency matrix of the $N$ adjacency matrices, $\sE{\bA}$.
\subsection{Inhomogeneous \ER Graphs}
We first consider the probability space $\gnP$ formed by inhomogeneous \ER random graphs \citep{bollobas07}, defined on
the vertex set $\{1,\ldots,n \}$, where a graph $G$ with adjacency matrix $\bA$ has probability,
\begin{equation}
  \prob{\bA} = \prod_{1 \le 1 < j \le n} \left [p_{ij} \right]^{a_{ij}} \left [1- p_{ij} \right]^{1 - a_{ij}}.
  \label{laproba}
\end{equation}
The $n \times n$ matrix $\bP = \left [p_{ij}\right]$ determines the edge probabilities $0
\le p_{ij} \le 1$, with $p_{ii} = 0$. In the case of a sample of inhomogeneous \ER random
graphs, all generated from the same probability matrix
$\bP$, one can easily compute the sample \fr mean graph, $\EG$. Indeed, the adjacency matrix $\EA$ of
\begin{algorithm}[H]
  \caption{Inhomogeneous \ER Training Data Generation}\label{alg1}
  \begin{algorithmic}
    \For{i in 1:36}
    \State a = uniform(0.5,5)
    \State b = uniform(0.5,5)
    \For{j in 1:100}
    \State $\bP$ = Beta(a,b)
    \For{k in 1:10}
    \State sample[k] = E.R.($\bP$)
    \EndFor
    \State $\bA_I = \sum_1^{10}\text{sample}$
    \State $\sE{\bA} = \text{KnownMean}(\bP)$
    \EndFor
    \State InputList.append($\bA_I$) 
    \State TargetList.append($\sE{\bA}$)
    \EndFor
  \end{algorithmic}
\end{algorithm}
\noindent $\EG$ is given by
\citep{meyer22b}
\begin{equation}
  \EA_{ij} = \begin{cases} 
    1 & \sE{\bA}_{ij} > \frac{1}{2}, \\
    0 & \text{otherwise}. 
  \end{cases}
  \label{FMIER}
\end{equation}
We varied the edge probability matrix, $\bP$. For each simulation, $\bP$ was chosen randomly using independent (up to
symmetry) beta random variables, $p_{ij} \sim \text{beta} (a, b)$. The parameters $a$ and $b$ were sampled uniformly
on $[0.5, 5]$. Alg.~\ref{alg1} provides a description of this process.
\subsection{Stochastic Block Model}
One important property of real world networks is community structure. Vertices form
densely connected communities, with the connection between communities being sparse, or
non-existent. This motivates the use of the stochastic blockmodel. In this model, the
vertex set can be partitioned into non-overlapping sets referred to as ``communities''.
Each edge $e = (i,j)$ exists independently with probability $p$ if $i$ and $j$ are in the
same community, and $q$ if $i$ and $j$ are in distinct communities.

The training dataset is formed of $m/2$ graph with two communities of size $n/2=14$, and $m/2$ graphs with three
communities with sizes 10, 10, and 8. We then sample the edge probabilities, $\mathbf{p}$ and $\mathbf{q}$ uniformly on
$[0.5, 0.9]$ and on $[0.1, 0.5]$ respectively. We restrict our samples to only connected graphs. Alg.~\ref{alg2}
provides a description of this process.
\begin{algorithm}[H]
  \caption{Stochastic Block Model Training Data Generation}\label{alg2}
  \begin{algorithmic}
    \Require BlockSize2 = $\left[14,14\right]$
    \Require BlockSize3 = $\left[10,10,8\right]$
    \For{$i$ in 1:36}
    \State $\mathbf{p}$ = uniform(0.5,0.9)
    \State $\mathbf{q}$ = uniform(0.01,0.5)
    \If{$i < 19$}
    \For{j in 1:100}
    \For{k in 1:10}
    \State $\mathcal{A}[k]$ = S.B.(p,q, BlockSize2)
    \EndFor
    \State $\bA_I = \sum_1^{10}\mathcal{A}$
    \State $\sE{\bA} = \argmin{G\in \mathcal{A}} \frac{1}{10}\sum_{k=1}^{10} d_A^2(G,\Gk)$
    \EndFor
    \ElsIf{$i \geq 19$}
    \For{j in 1:100}
    \For{k in 1:10}
    \State $\mathcal{A}[k]$ = S.B.(p,q, BlockSize3)
    \EndFor
    \State $\bA_I = \sum_1^{10}\mathcal{A}$
    \State $\sE{\bA} = \argmin{G\in \mathcal{A}} \frac{1}{10}\sum_{k=1}^{10} d_A^2(G,\Gk)$
    \EndFor
    \EndIf
    \State InputList.append($\bA_I$) 
    \State TargetList.append($\sE{\bA}$)
    \EndFor
  \end{algorithmic}
\end{algorithm}
\noindent 
\subsection{Preferential Attachment Graphs}
Another frequently studied feature for real world graphs is the degree distribution. Real world graphs, such as social
and computer networks, have a power-law tail, $\pr(d) \propto d^{\gamma}$ where $\gamma \in \left[2,3\right]$. These
distributions are referred to as ``scale-free" \citep{Barabasi1999}.

The preferential attachment model is a random graph model that fits the characteristics of ``scale-free". The model has
two parameters, $l$ (density of the graph) and $n$ (size of the graph). We begin by initializing a star graph with $l+1$
vertices, with vertex $l+1$ having degree $l$ and all others having degree $1$. Then, for each $l+1<i\leq n$, we add a
vertex, and randomly attach it to $l$ vertices already present in the graph, where the probability of $i$ attaching to
$v$ is proportional to to the degree of $v$. We stop once the graph contains $n$ vertices.

The training set includes $m$ graphs  for a variety of $l$ values (see  Alg.~\ref{alg3}).
\begin{algorithm}[H]
  \caption{Preferential Attachment Data Generation}\label{alg3}
  \begin{algorithmic}
    \Require $l = \left[5, 7, 10, 12, 15, 17, 20, 22, 25\right]$
    \For{$i \in l$}
    \For{$j$ in 1:400}
    \For{$k$ in 1:10}
    \State $\mathcal{A}[k]$ = P.A.(l,n=28)
    \EndFor
    \State $\bA_I = \sum_1^{10} \mathcal{A}$
    \State $\sE{\bA} = \argmin{G\in \mathcal{A}} \frac{1}{10}\sum_{k=1}^{10} d_{\mathcal{L}}^2(G,\Gk)$
    \EndFor
    \State InputList.append($\bA_I$) 
    \State TargetList.append($\sE{\bA}$)
    \EndFor
  \end{algorithmic}
\end{algorithm}
\noindent
\subsection{Training}
We train four distinct versions of the same exact model described in Section~\ref{CNN} using four different training
sets. We call each version of the algorithm with a different name that encodes the name of the graph ensembles used to
train the model: IER-Unet, SBM-Unet, BA-UNet. We also consider a model, Gen-Unet, that is trained using a training set
formed by the union of the three previous training sets.

We employ batch Adam optimization coupled with a binary cross entropy as our loss in order to train our models. Binary
cross entropy is a natural choice for our loss function as it is widely used in binary classification tasks. We refer
the reader to \citep{kingma15} for more details pertaining to Adam.
\section{Experiments
  \label{experiments-section}}
We validate our approach with a wide variety of experiments using the three random graph ensembles. For each experiment,
we report several statistics that characterize the precision of our estimate of the sample \fr mean graph.

Our test data is composed of $l= 3\times 900$ adjacency matrices from each of the three random graph ensembles: \ER,
stochastic block model, and preferential attachment. In the next section, we describe the methodology that we use to
quantify the results of the experiments.
\subsection{Evaluation protocol}
In the following, we describe the gauges that we use to quantify the error between the true and the estimated sample
\fr mean graph. Our comparison combines two different scales: the large to medium scale connectivity quantified  by the
largest eigenvalues of the adjacency matrix, and the fine scale connectivity, formed by the ego-net quantified by the
degree distribution.
\subsubsection*{The spectral differences.}
\begin{definition}
  Given a sample of $N$ adjacency matrices, $\bA_1,\ldots,\bA_N$, and their vectors of ordered eigenvalues (see
  Def. \ref{spectrum}), $\blb(\bA_1), \ldots, \blb(\bA_N)$, the sample mean vector of eigenvalues is given by,
  \begin{equation}
    \sE{\blb} = \frac{1}{N}\sum_{k=1}^N \blb(\bA_k).
  \end{equation}
  We define the vector of eigenvalues of the sample \fr mean as follows,
  \begin{equation}
    \lFM = \blb(\EA).
  \end{equation}
  Finally, we denote by $\olF$, the spectrum of our estimate, $\oEA$ of the sample \fr mean,
  \begin{equation}
    \olF = \blb(\oEA).
  \end{equation}
\end{definition}
To simplify the exposition, and alleviate the complexity of the notations, we remove the notation $[\bA]$ in $\sE{\blb},
\lFM$, and $\olF$, which indicate the dependency of the sample mean on the adjacency matrices in the sample,
$\bA_1,\ldots,\bA_N$.

We expect that $\sE{\blb}$ be close to $\lFM$, since as shown in \citep{ferguson22}, the sample mean spectrum is
asymptotically close to the spectrum of the sample \fr mean, in the limit of large graph size. In order to compare
$\lFM$ and our estimate $\olF$, we define the vector of absolute differences
($\bDlt_{\blb}$).
\begin{definition}
  Let $\bA_1$ and $\bA_2$ be the adjacency matrices of two graphs. Let  $\blb(\bA_1)$ and $\blb(\bA_2)$ be their
  respective eigenvalues vectors (with entries sorted by their magnitude). We denote by 
  \begin{equation}
    \bDlt_{\blb}(\bA_1, \bA_2) =
    \begin{bmatrix}
      \lvert \lambda_1(\bA_1) - \lambda_1(\bA_2) \rvert & \ldots & \lvert \lambda_n(\bA_1) - \lambda_n(\bA_2)\rvert
    \end{bmatrix},
    \label{ADE}
  \end{equation}
  the vector of absolute differences between the eigenvalues of $\bA_1$ and $\bA_2$.
\end{definition}
Because the eigenvalues of the adjacency matrix are not normalized, we introduce the vector of relative
difference between two vectors of eigenvalues with the following definition.
\begin{definition}
  Let $\bA_1$ and $\bA_2$ be the adjacency matrices of two graphs. Let  $\blb(\bA_1)$ and $\blb(\bA_2)$ be their
  respective eigenvalues vectors (with entries sorted by their magnitude). We denote by 
  \begin{equation}
    \bDlt_{\blb}^p(\bA_1, \bA_2) =
    \begin{bmatrix}
      \displaystyle
      \frac{\lvert \lambda_1(\bA_1) -  \lambda_1(\bA_2) \rvert}
           {\lvert \lambda_1(\bA_2) \rvert} &
           \dots &
           \displaystyle \frac{ \lvert \lambda_n(\bA_1) -  \lambda_n(\bA_2)
             \rvert}{ \lvert \lambda_1(\bA_2)\rvert}
   \end{bmatrix},
    \label{PADE}
  \end{equation}
  the vector of relative difference per eigenvalue.
\end{definition}
In this paper, we are concerned with the comparison of the eigenvalues of $\oEA$ and these of $\EA$. We compute the
vectors of (absolute and relative) differences, $\bDlt$ and $\bDlt^p$. In order to alleviate the complexity of the
notations, we remove the notation $[\bA]$, which indicate the dependency of the sample mean on the adjacency matrices in
the sample, $\bA_1,\ldots,\bA_N$, in these differences. We therefore define,
\begin{equation}
  \bDlt_{\blb} (\sfm, \ofm) \eqdef \bDlt_{\blb}(\EA,\oEA) \quad\text{and}\quad 
  \bDlt^p_{\blb} (\sfm, \ofm) \eqdef \bDlt^p_{\blb}(\EA,\oEA).
\end{equation}

For the evaluation of each model, we repeat our testing experiments $l=900$ times for each of three models. We therefore
define the average (absolute or relative) vector of eigenvalues -- computed over the different estimates of the
eigenvalues for the same model.
\begin{definition}
  Given a set of spectral differences $\left\lbrace \bDlt^{(k)}_{\blb}\right\rbrace_{k=1}^{s}$, computed
  over $l$ independent trials, the average absolute difference per eigenvalue is the $n$-dimensional vector given by,
  \begin{equation}
    \overline{\bDlt_{\blb}} = \frac{1}{l} \sum_{k=1}^l \bDlt^{(k)}_{\blb}.
    \label{AADE} 
  \end{equation}
\end{definition}
\begin{definition}
  Given a set of spectral relative differences $\left\lbrace [\bDlt_{\blb}^p]^{(k)} \right\rbrace_{k=1}^{s}$, computed
  over $s$ independent trials, the average relative difference per eigenvalue is the $n$-dimensional vector given by,
  \begin{equation}
    \avgDelP = \frac{1}{l} \sum_{k=1}^l  [\bDlt_{\blb}^p]^{(k)}.
    \label{APADE}
  \end{equation}
\end{definition}


\subsubsection{The distance between the degree distributions.}

The fine scale statistics, such as the degree distribution, provides a ``window'' on patterns of connectivity that
happen at a fine local scale (the egonet). This small scale statistic is worthy of attention because large communities
in real networks can be decomposed into smaller scale communities, which become more and more connected, as they become
smaller \citep{leskovec09}. The degree distribution quantifies the finest scale ``atomic'' communities that cannot be
further reduced.

Given a graph $G=(V,E)$ of size $n$, we use the histogram $f_n$ of degrees to characterize the degree distribution, defined as
follows
\begin{equation}
  f_n(k) = \frac{n_k}{n}
\end{equation}
where $n_k$ is the number of vertices in $V$ with degree $k$. To compare two degree histograms, $f_n$ and $g_n$, we use
the Kullback-Leibler divergence \citep{kullback51} given by
  \begin{equation}
    D_{KL}(f_n,g_n) = \sum_{k=1}^nf_n(k)\log\left(\frac{f_n(k)}{g_n(k)}\right).   \label{KL}
  \end{equation}
The smaller the value of $D_{KL}(f_n,g_n)$ the closer the two distributions are to one another
\subsection{A pedestrian tour of the testing process.}
To help provide some understanding into the testing process, we provide a tour of the testing process (for a specific
model: SBM-Unet). Other models are evaluated in the same manner.

After having trained the convolutional neural network, we test the algorithm with a sample of adjacency matrices $\bA_1,
\dots, \bA_N$. We first compute the sample mean adjacency matrix $\sE{\bA}$, which is the input to our algorithm. Our
algorithm returns $\ofm$, our estimate of the sample \fr mean. We then compute the eigenvalues $\blb(\sfm)$ and
$\blb(\ofm)$, and the vectors of errors in the eigenvalues, $ \bDlt (\sfm, \ofm)$, and their relative versions $
\bDlt^{p} (\sfm, \ofm)$.

\begin{figure}[H]
  \centerline{
    \includegraphics[width=0.5\textwidth]{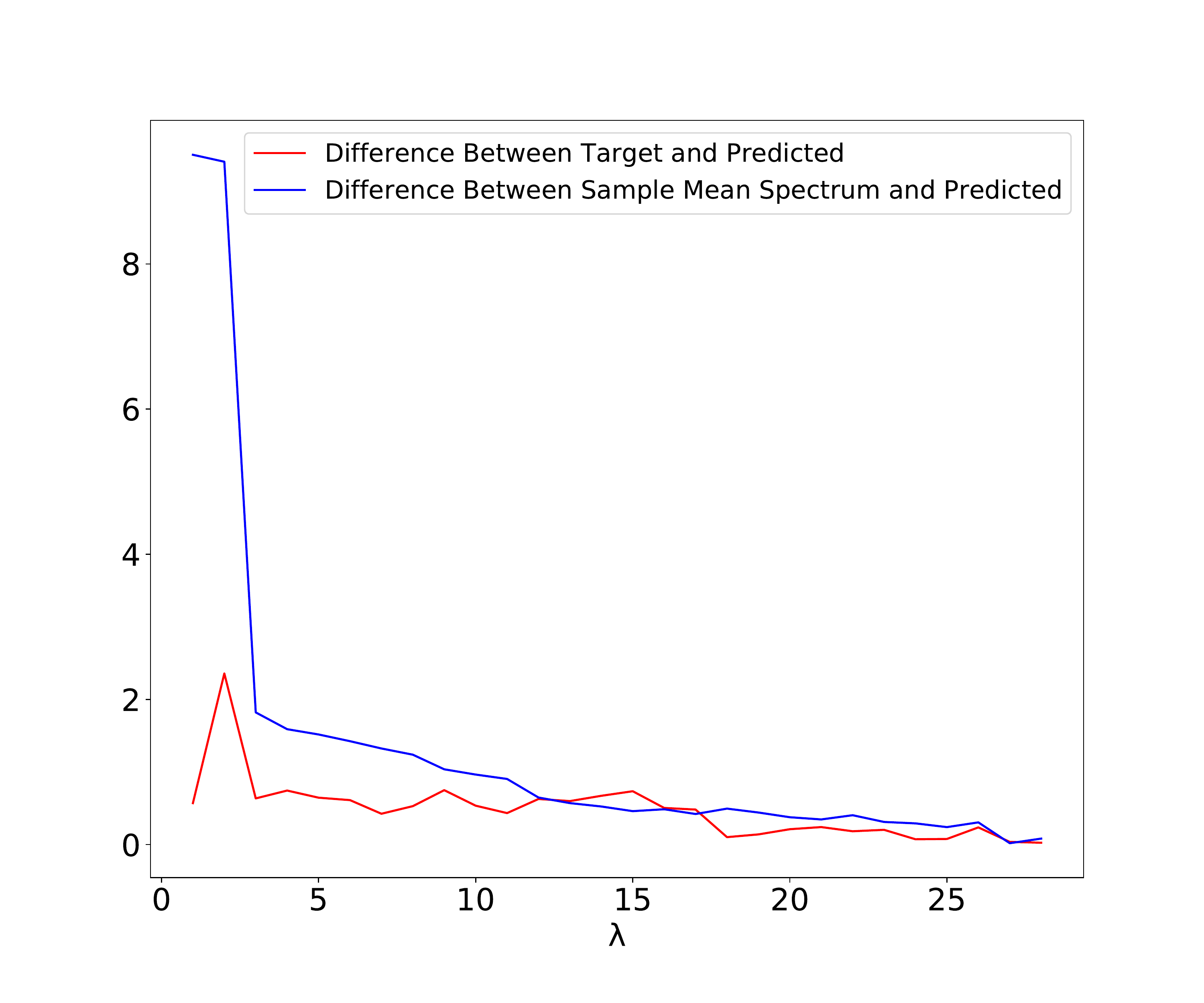}
    \hfill
    \includegraphics[width=0.5\textwidth]{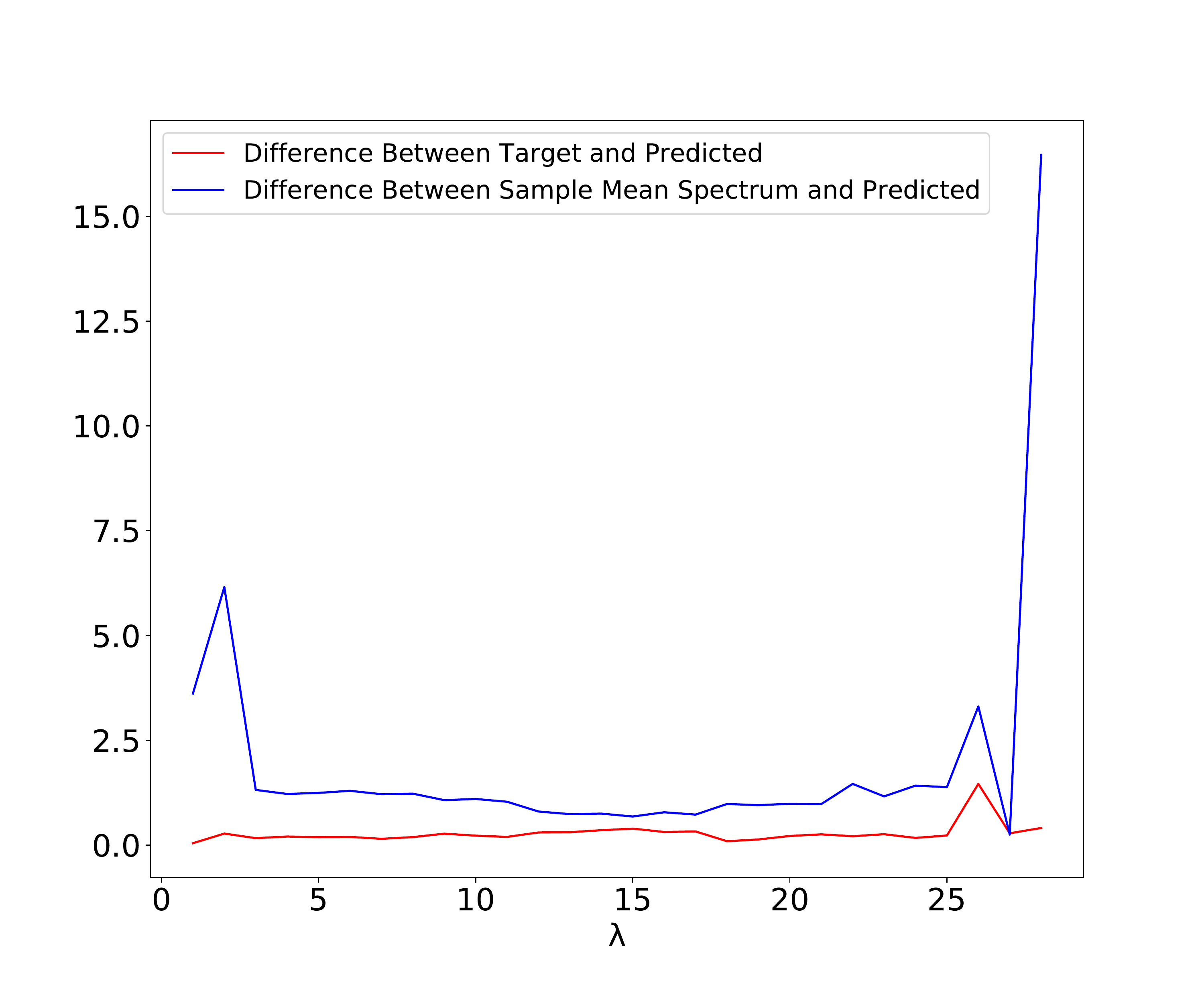}
  }
  \caption{The vector of eigenvalues errors $\bDlt(\sfm,\ofm)$ (in red) and $\bDlt(\oEA, \sE{\blb})$
    (in blue). Left: absolute error; right: relative error.}
  \label{fig3}
\end{figure}

Fig.~\ref{fig3}-left displays (in red) the 28 eigenvalue errors, $\bDlt (\sfm,\ofm)$, while Fig.~\ref{fig3}-right
displays the relative errors $\bDlt^p (\sfm,\ofm)$. In addition, we also display (in blue) the error between sample mean
spectrum, $\sE{\blb}$, and the spectrum of our estimate of the \fr mean, $\blb(\ofm)$.

The visual inspection of Fig.~\ref{fig3} suggests that the spectral differences, $\bDlt_{\blb}(\sfm,\ofm)$ (red curve)
between the predicted adjacency matrix $\oEA$ and the true \fr mean $\EA$,  is maximal for
$\lambda_2$. Interestingly, the sample mean spectrum $\blb(\sE{\bA})$ remains very close to $\blb(\oEA)$  (except for the
first eigenvalue).

\begin{figure}[H]
  \centerline{
    \includegraphics[width=.7\textwidth]{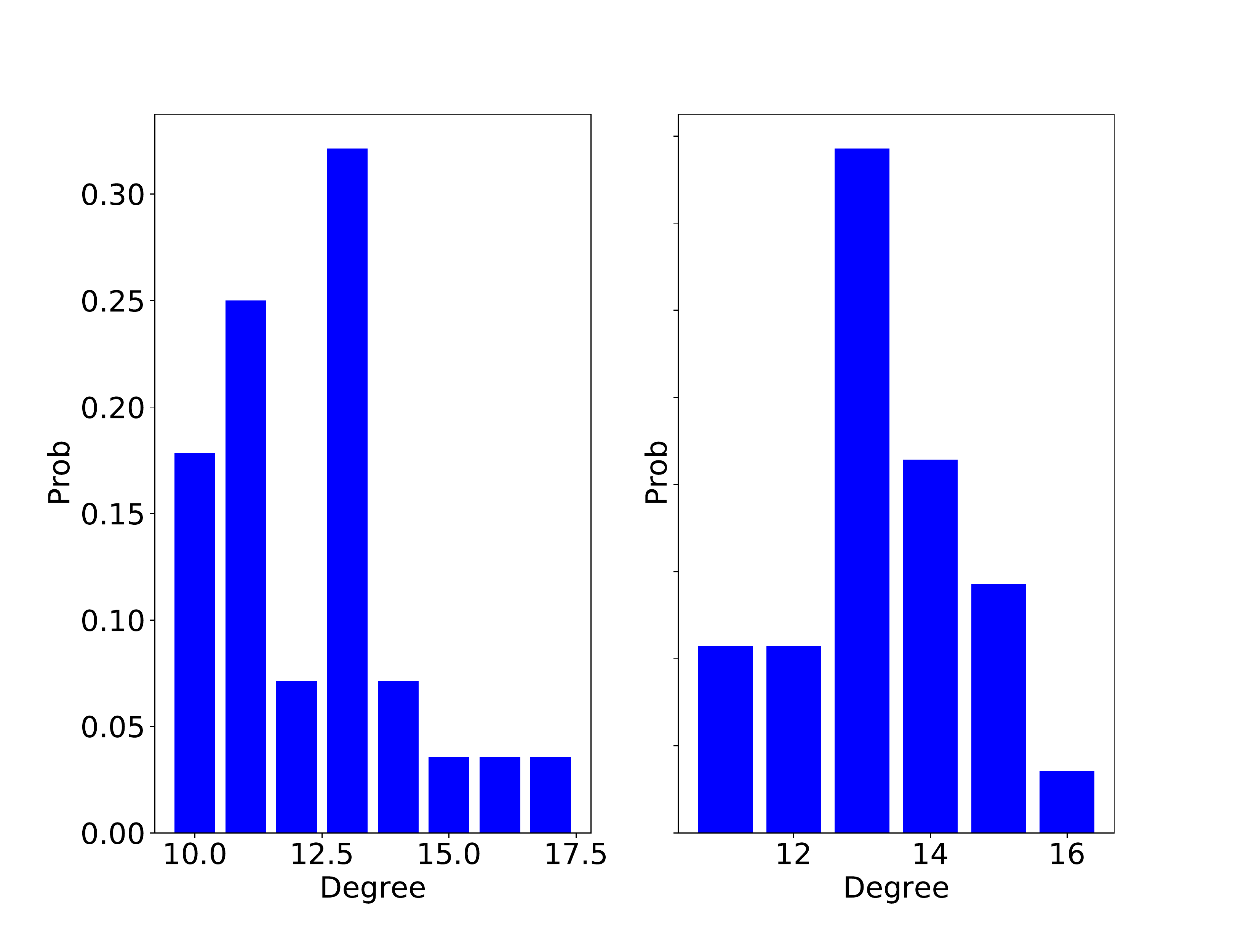}
  }
  \caption{Degree distribution of the true sample \fr mean graph, $\sfm[G]$, (left) and of the predicted sample \fr mean
    graph, $\ofm[G]$ (right).
    }
  \label{fig4}
\end{figure}

Next, we calculate the degree distributions both for the predicted sample \fr mean graph, $\ofm[G]$, as well as the true
sample \fr mean graph, $\sfm[G]$ (see Fig.~\ref{fig4}).

\noindent

We notice in Fig.~\ref{fig4} that while the distributions appear to be fairly different, SBM-Unet is able to accurately
capture the peak corresponding to a degree of 13 in the true sample \fr mean graph, $\sfm[G]$. The KL divergence between the
two distribution is approximately $3.8$.
\section{Results and Discussion}

In general, we find that at least one of our models performs better than the naive $\nfm$ both in terms of $\avgDel
(\nfm,\sfm)$ and $\avgDelP(\nfm,\sfm)$. This suggests that our models are able to better predict large scale structures
of the sample \fr mean than the naive model. The spectral error between our prediction and the naive model is
significantly smaller for the first eigenvalue, and remains lower up to the fifth or sixth eigenvalue.

When comparing the eigenvalues of the predicted \fr mean, $\blb(\ofm)$, and the sample mean vector of eigenvalues
$\sE{\blb}$, we find that at on average the eigenvalues from the \fr means computed with our models are closer to the
sample mean eigenvalues than the eigenvalues predicted by the naive model $\blb(\nfm{\bA})$. 

Finally, with regard to the degree distributions we find it is often the case that our models produce degree
distributions that are closer in terms of KL divergence to the true sample \fr mean graph, $\sfm[G]$ than to the naive $\nfm[G]$.
\subsection{Inhomogeneous \ER Testing Data}
\begin{figure}[H]
  \centerline{
    \includegraphics[width=0.5\textwidth]{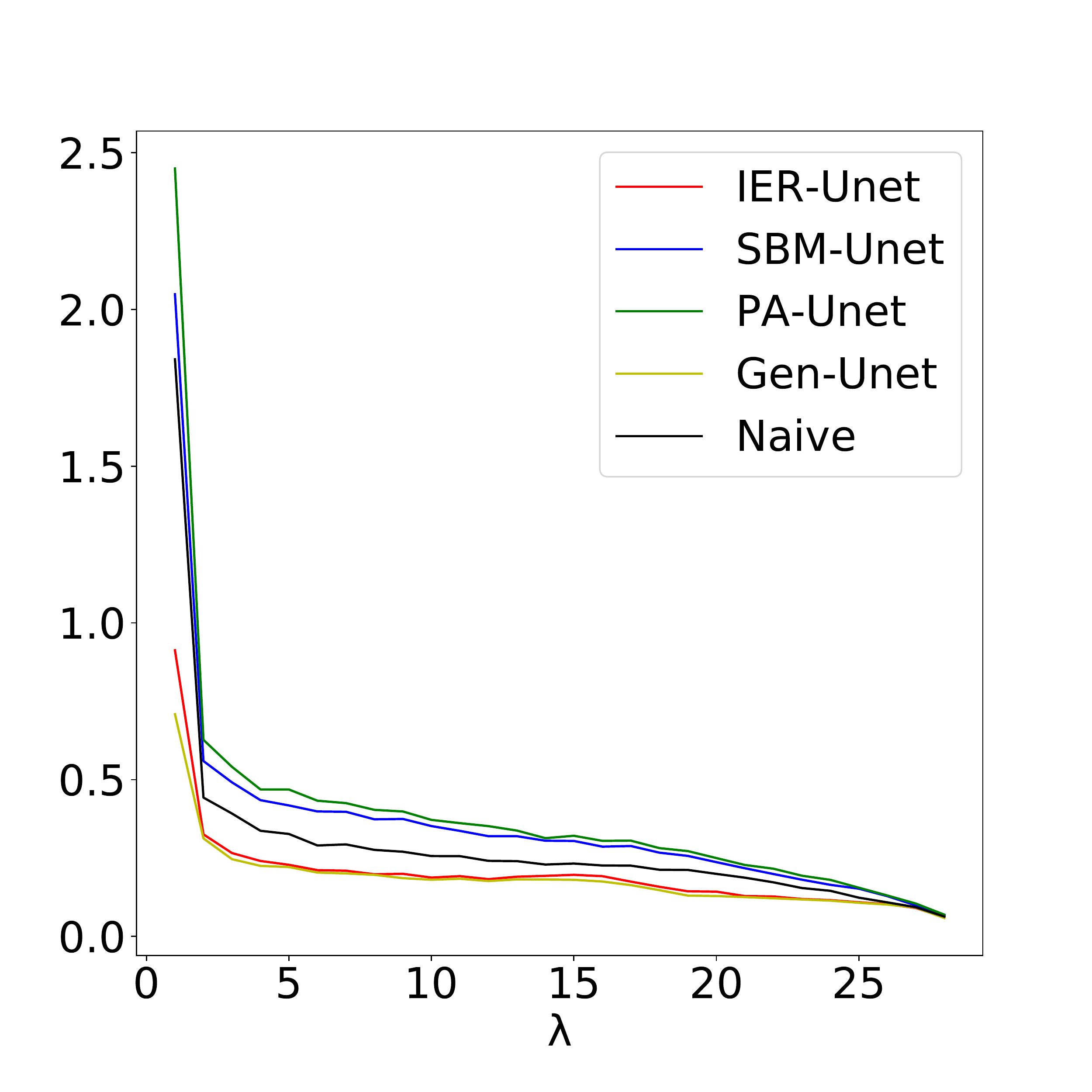}
    \hfill
    \includegraphics[width=0.5\textwidth]{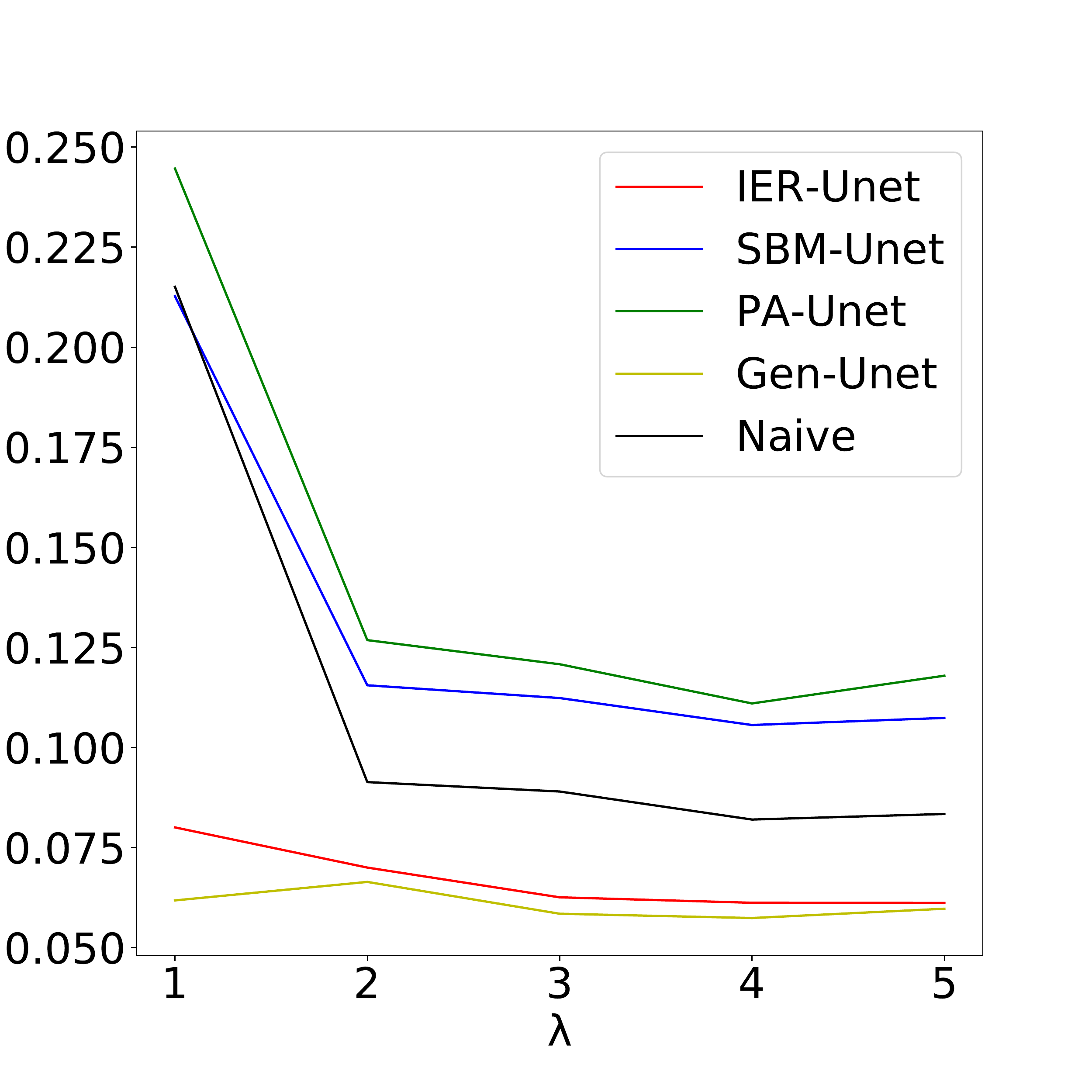}
  }
  \caption{Inhomogeneous \ER testing data: error between the eigenvalues of the true and predicted \fr mean for each model; absolute error
    $\avgDel(\sfm,\ofm)$ for the first 25 eigenvalues (left); relative error for the first five eigenvalues,
    $\avgDelP(\sfm,\ofm)$ (right).
    \label{fig5}}
\end{figure}

When the four models were evaluated using inhomogeneous \ER random graphs, IER-Unet and Gen-Unet resulted in the lowest
spectral approximation error (see Fig.~\ref{fig5}). In contrast, both the IER-Unet and Gen-Unet had virtually identical
$\avgDel(\sfm,\ofm)$ and $\avgDelP(\sfm,\ofm)$, and performed worse than the naive model.

Fig.~\ref{fig5} provides a plot of the entries of the spectral differences $\bDlt_{\blb}(\sfm,\ofm)$ and
$\bDlt_{\blb}^p(\sfm,\ofm)$. Tables~\ref{tbl1} and ~\ref{tbl2} provide some basic statistics for both
\begin{figure}[H]
  \centerline{
    \includegraphics[width=0.5\textwidth]{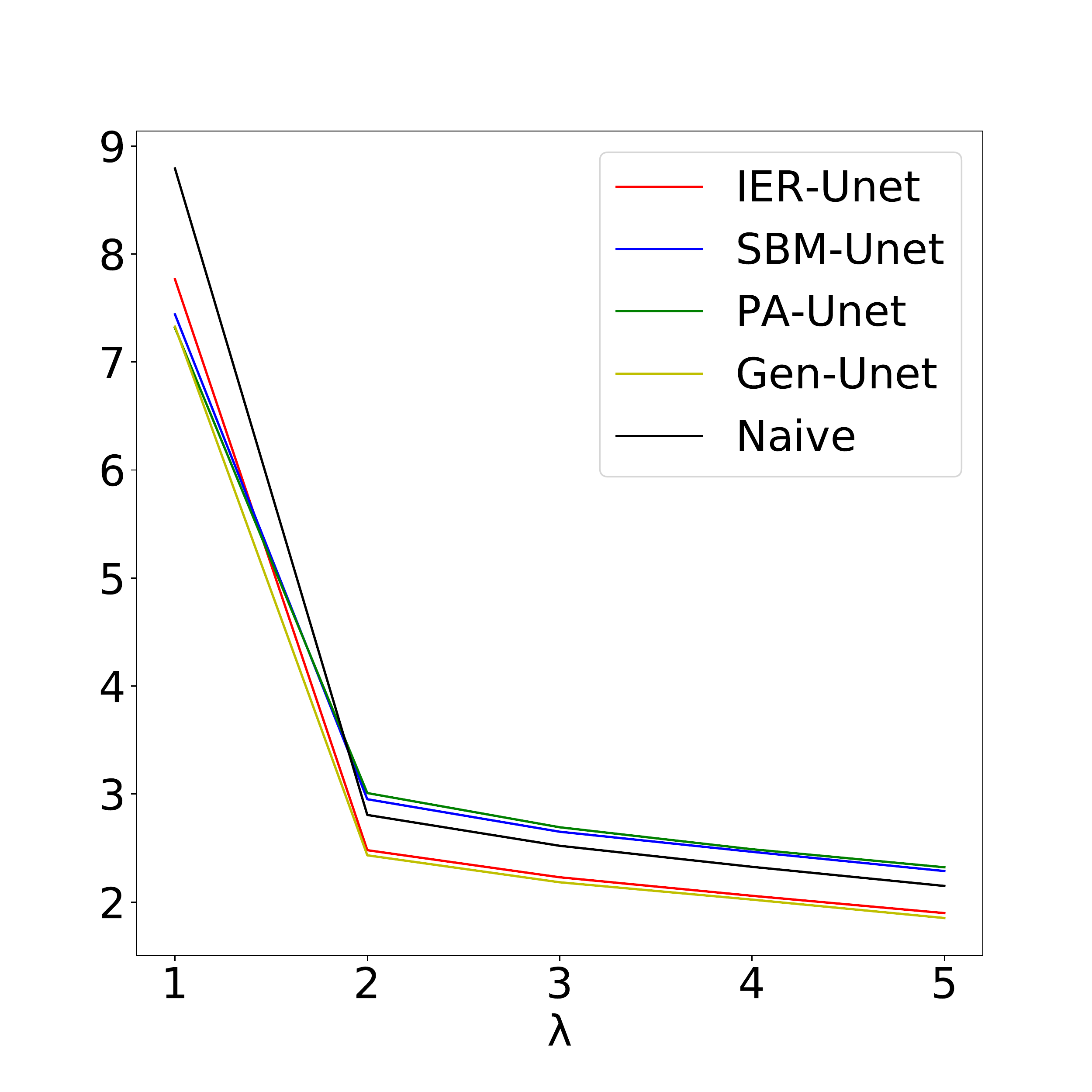}
    \hfill
    \includegraphics[width=0.5\textwidth]{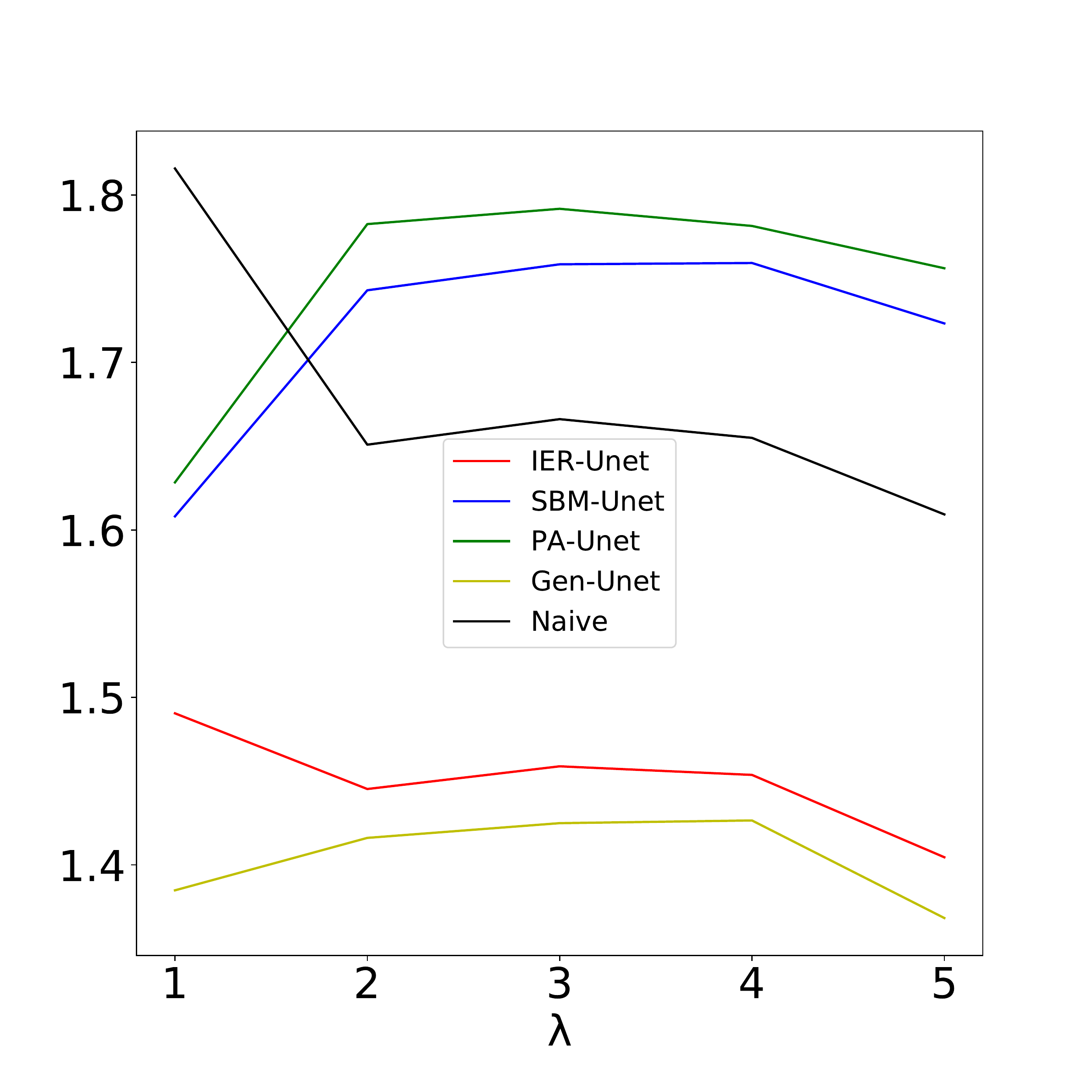}
  }
  \caption{Inhomogeneous \ER testing data: spectral difference between the eigenvalues of the predicted \fr mean and the sample mean eigenvalues;
    absolute difference $\avgDel(\blb(\ofm), \sE{\blb} )$ for the first 5 eigenvalues (left); relative difference for the first
    five eigenvalues, $\avgDelP(\blb(\ofm), \sE{\blb} )$ (right).
    \label{fig6}}
\end{figure}

$\avgDel(\sfm,\ofm)$ and $\avgDelP(\sfm,\ofm)$, including the maxima and minima of the entries, as well as the index of
the corresponding minimum or maximum.

The comparison of the sample mean spectrum, $\sE{\blb}$ and the spectrum, $\blb(\ofm)$ (see Fig.~\ref{fig6}) reveals
that all models perform relatively similarly in terms of $\avgDel(\blb(\ofm),\sE{\blb})$ and
$\avgDelP(\blb(\ofm),\sE{\blb})$. Overall, all our models achieve a smaller spectral error than the naive model. This
results implies that our models, regardless of the training data, are able to capture the connectivity information that
controls the eigenvalues of the adjacency matrix, irrespective of the training data.
This result is not surprising, since Gen-Unet and IER-Unet were both presented with inhomogeneous \ER training data, and
therefore we expected them to be able to predict the \ER graphs with high precision.
\begin{table}[H]
  \begin{center}
      \caption{$\avgDel(\sfm,\ofm)$ statistics for \ER generated testing
        data}\label{tbl1}
      \begin{tabular}{@{}lrcrc@{}} 
        \toprule 
        Model & $\max \avgDel$ & $\mspace{24mu}\lambda_{\bDlt}^{\text{max}}$ & 
        $\mspace{24mu}\min \avgDel$ & $\mspace{24mu}\lambda_{\bDlt}^{\text{min}}$\\
        \midrule
        IER-Unet  & 0.912596 & 1 & 0.060550 & 28 \\
        SBM-Unet  & 2.048245 & 1 & 0.066471 & 28 \\
        PA-Unet  & 2.449364 & 1 & 0.069327 & 28 \\
        Gen-Unet  & 0.708872 & 1 & 0.058604 & 28 \\
        Naive  & 1.840880 & 1 & 0.063173 & 28 \\
        \botrule
    \end{tabular}
  \end{center}
\end{table}
\begin{table}[H]
  \caption{$\avgDelP(\sfm,\ofm)$ statistics for \ER generated testing data, $\max \avgDelP$ is over the first 10 eigenvalues only.}
  \label{tbl2}
  \begin{center}
    \begin{tabular}{lrcrc}
      \toprule 
      Model & $\max \avgDelP$ & $\mspace{24mu}\lambda_{\bDlt^p}^{\text{max}}$ & 
      $\mspace{24mu}\min \avgDelP$ & $\mspace{24mu}\lambda_{\bDlt^p}^{\text{min}}$\\
      \midrule
      IER-Unet   & $0.080035$ & 1 & 0.060586 & 6 \\
      SBM-Unet  & $0.212734$ & 1 & 0.105618 & 4 \\
      PA-Unet   & $0.244647$ & 1 & 0.111006 & 4 \\
      Gen-Unet  & $0.070039$ & 10 & 0.057397 & 4 \\
      Naive  & $0.215085$ & 1 & 0.078074 & 6 \\
      \botrule
    \end{tabular}
  \end{center}
\end{table}
\begin{table}[H]
  \caption{KL divergence for the inhomogeneous \ER testing data}
  \label{tbl3}
  \begin{center}
    \begin{tabular}{lrr}
      \toprule 
      Model & Mean & Variance\\
      \midrule
      IER-Unet   & 0.6066 & 0.0737 \\
      SBM-Unet  & 0.9588 & 0.2772 \\
      PA-Unet   & 0.9899 & 0.0488 \\
      Gen-Unet  & 0.4133 & 0.04877 \\
      Naive  & 0.4610 & 0.0675 \\
      \botrule
    \end{tabular}
  \end{center}
\end{table}
When comparing the degree distributions of the true \fr mean graph, $\sfm[G]$, and that of the predicted sample \fr
mean, $\ofm[G]$, we again confirm that both IER-Unet and Gen-Unet have lowest average KL divergence (see Table~\ref{tbl3}).

\subsection{Stochastic Block Model Testing Data}
\begin{figure}[H]
  \centerline{
    \includegraphics[width=0.5\textwidth]{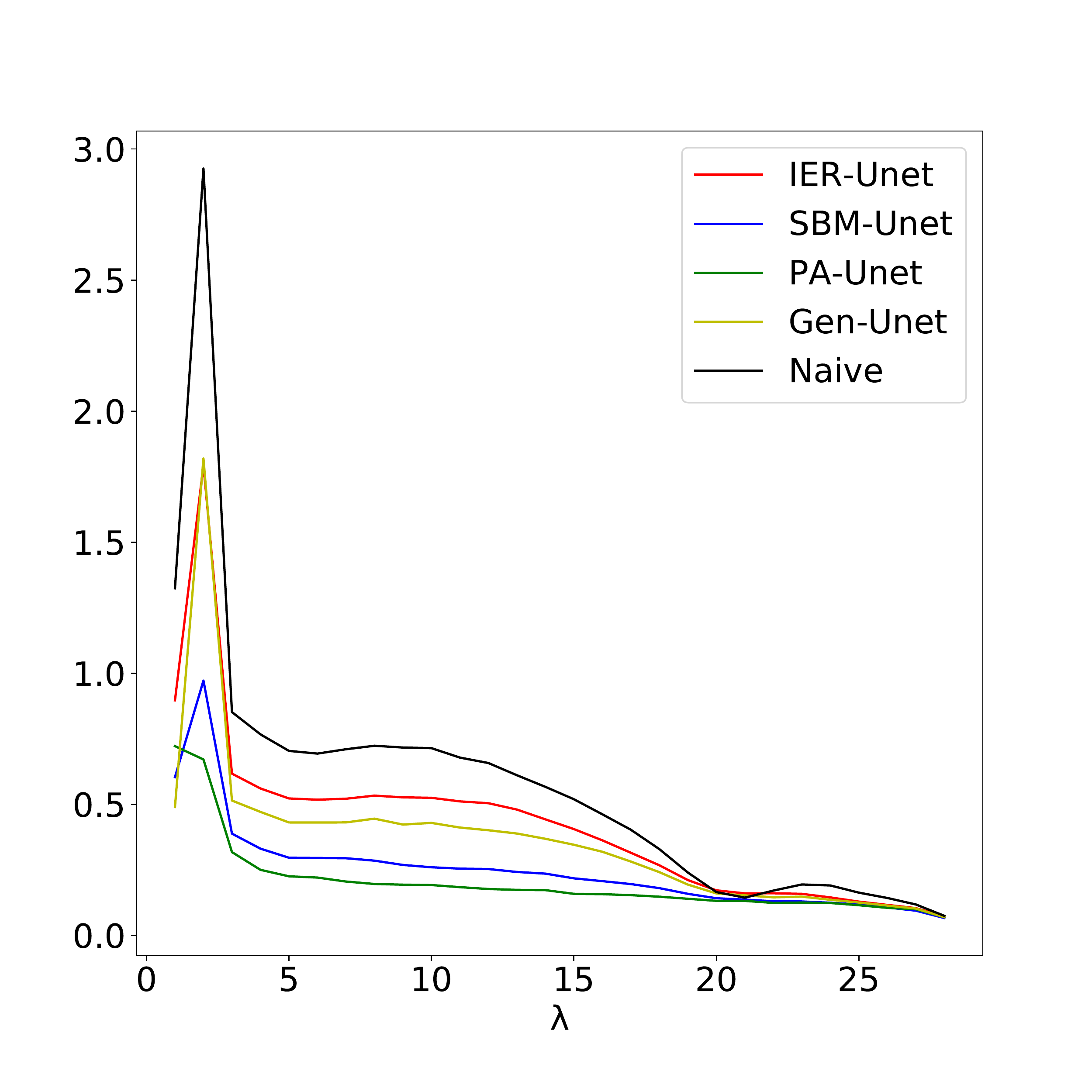}
    \hfill
    \includegraphics[width=0.5\textwidth]{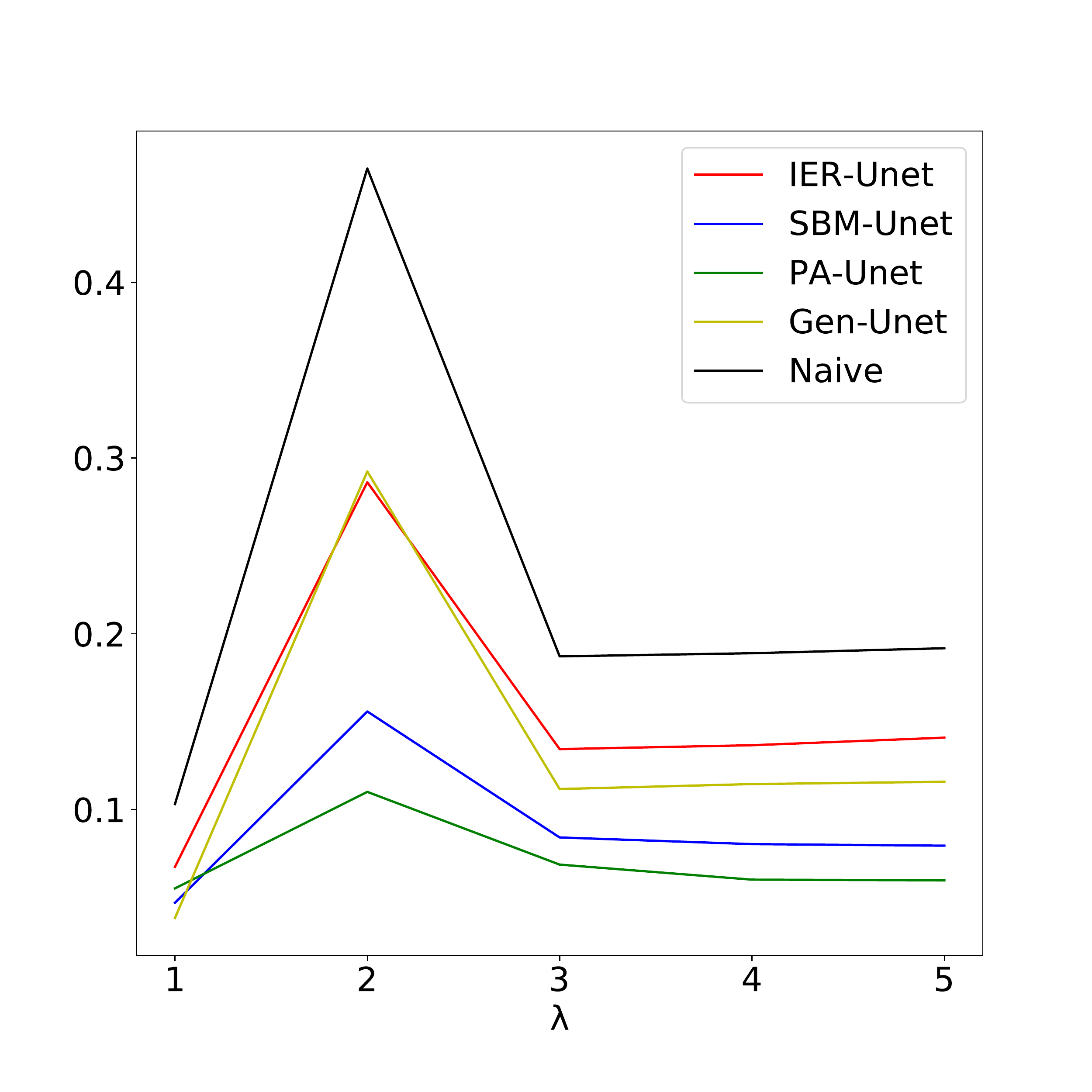}
  }
  \caption{Stochastic Block Model testing data: error between the eigenvalues of the true and predicted \fr mean for each model; absolute error
    $\avgDel(\sfm,\ofm)$ for the first 25 eigenvalues (left); relative error for the first five eigenvalues,
    $\avgDelP(\sfm,\ofm)$ (right).
    \label{fig7}}
\end{figure}

When evaluated using stochastic block model data, all of our models perform better than the naive model in terms of
$\avgDel(\sfm,\ofm)$ and $\avgDelP(\sfm,\ofm)$. Somewhat surprisingly, PA-Unet, performs the best out of all our models
followed closely by SBM-Unet (see Fig ~\ref{fig7}).

\begin{figure}[H]
  \centerline{
    \includegraphics[width=0.48\textwidth]{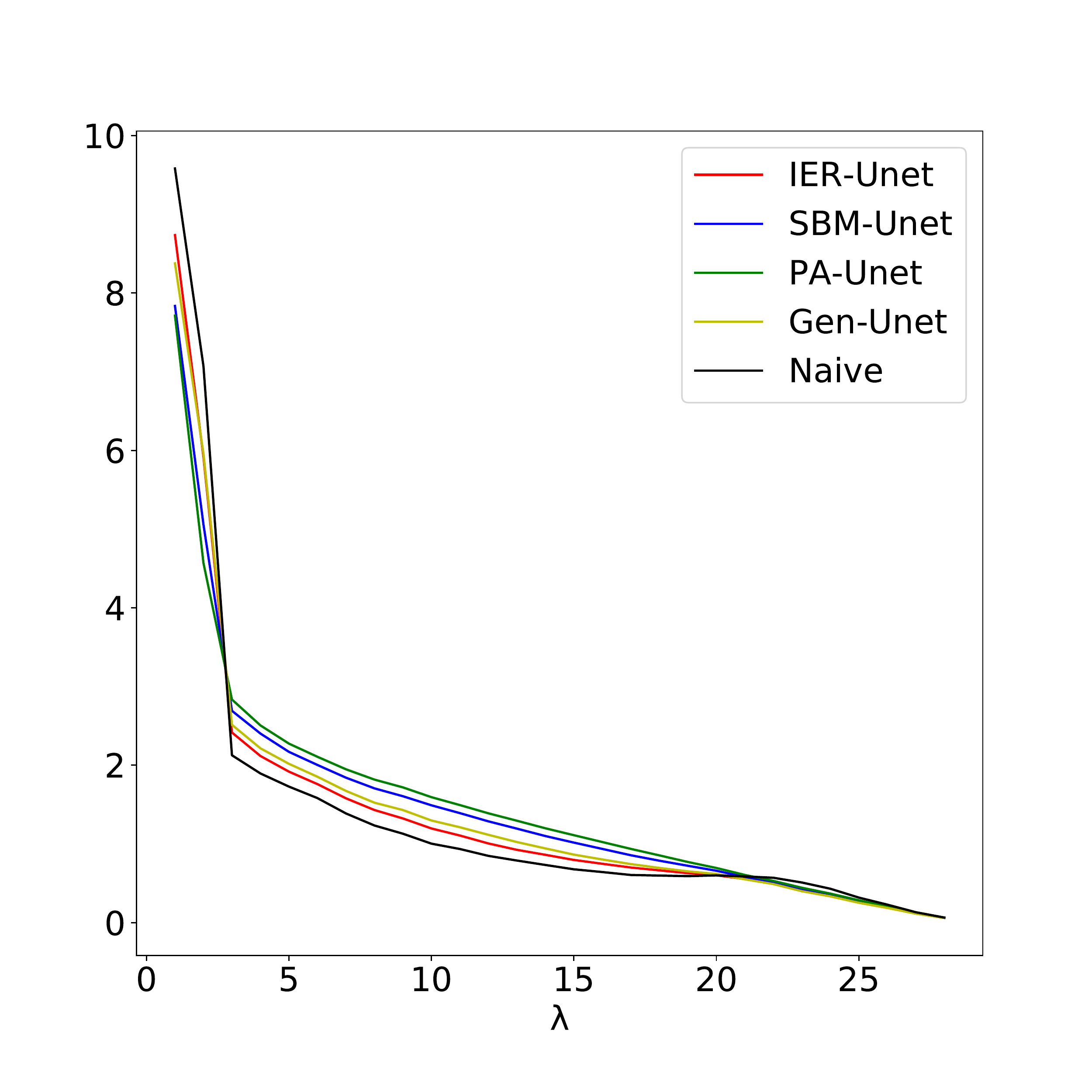}
    \hfill
    \includegraphics[width=0.47\textwidth]{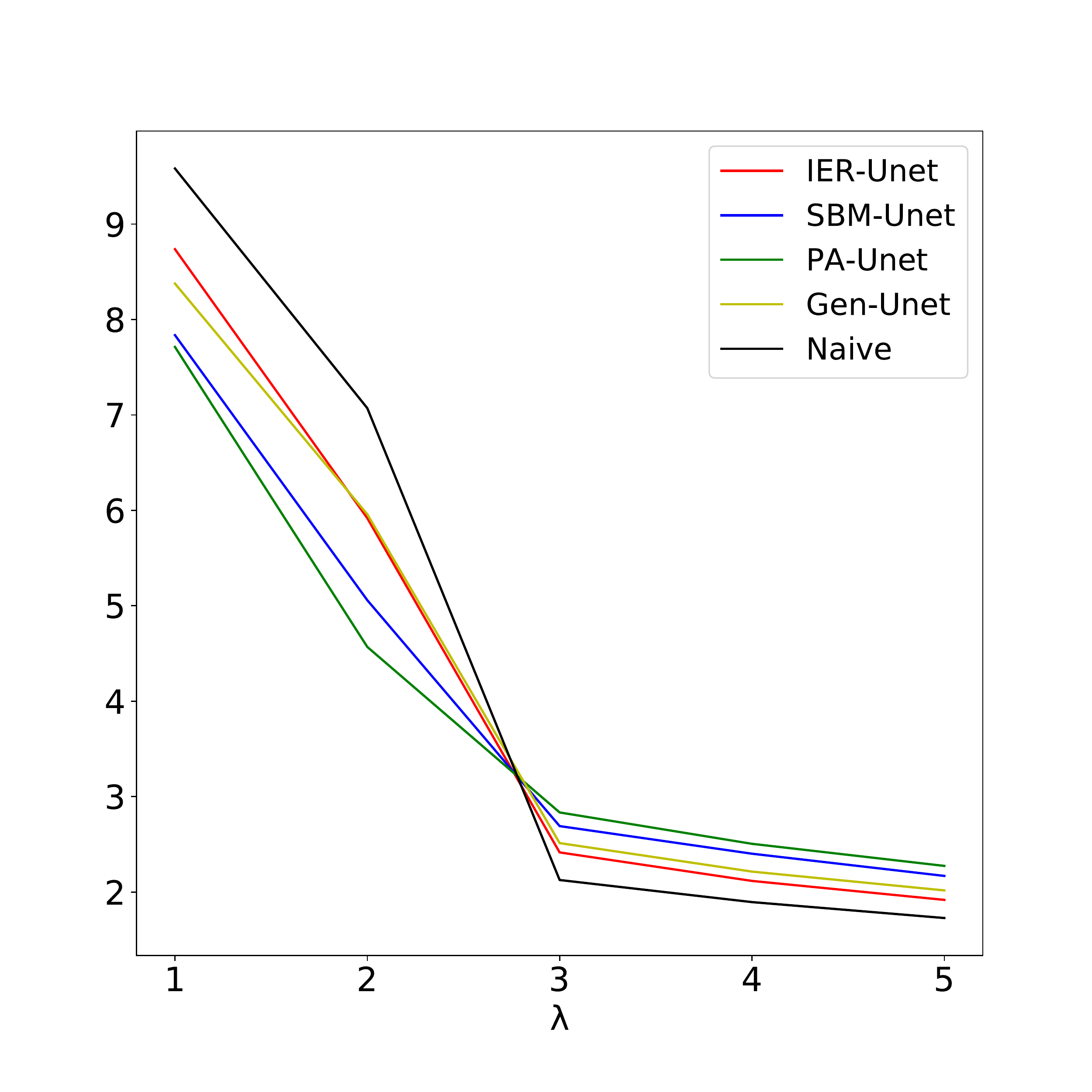}
  }
  \caption{Stochastic Block Model testing data: spectral difference between the eigenvalues of the predicted \fr mean
    and the sample mean eigenvalues; absolute difference $\avgDel(\blb(\ofm), \sE{\blb} )$ for the first 5 eigenvalues
    (left); relative difference for the first five eigenvalues, $\avgDelP(\blb(\ofm), \sE{\blb} )$ (right).
    \label{fig8}}

\end{figure}

The evaluation of the spectral difference between the eigenvalues of the predicted \fr mean and the sample mean
eigenvalues, $\avgDel(\blb(\ofm),\sE{\blb})$ and $\avgDelP(\blb(\ofm),\sE{\blb})$, indicate that on average, up until
the third eigenvalue, PA-Unet and SBM-Unet perform the best (see Fig.~\ref{fig8}). However, after the third eigenvalue
the naive model has the lowest values of $\avgDel(\blb(\ofm),\sE{\blb})$ and $\avgDelP(\blb(\ofm),\sE{\blb})$, although
the spectral difference does appear to converge towards zero as the eigenvalue increase for each model (see
Fig.~\ref{fig8}).

The change in model performance after the third eigenvalue suggests that our models are better predicting and
representing the salient structural features of the sample, however they may not be picking up other spectral
information as well as the naive model. Indeed, we know that the spectrum of the adjacency matrix of the SBM is composed
of a bulk centered around zero,
\begin{table}[H]
  \caption{KL divergence statistics for Stochastic Block Model generated testing data}
  \label{tbl4}
  \begin{center}
    \begin{tabular}{lrr}
      \toprule 
      Model & Mean & Variance\\
      \midrule
      IER-Unet   & 0.5034 & 0.0911 \\
      SBM-Unet  & 0.4366 & 0.0419 \\
      PA-Unet   & 0.5996 & 0.0488 \\
      Gen-Unet  & 0.3684 & 0.0244 \\
      Naive  & 0.4334 & 0.0476 \\
      \botrule
    \end{tabular}
  \end{center}
\end{table}
\noindent  and three positive eigenvalues -- each associated with a corresponding community. Our
models are therefore able to accurately predict the three dominant eigenvalues outside of the bulk, but miss the
eigenvalues in the bulk.

When comparing the degree distribution of the true sample \fr mean graph, $\sfm[G]$, to that of the predicted sample \fr
mean graph, $\ofm[G]$, we find that Gen-Unet achieves the lowest KL divergence (along with lower variance) (see
Table~\ref{tbl4}).
\subsection{Preferential Attachment Testing Data}
Preferential attachment models have long been favored for their similarities to real world networks. However as we have
discussed, less is known analytically about the eigenvalues of the preferential attachment model. We find that all of
our models perform better than the naive model, $\nfm$, in terms of $\avgDel(\sfm,\ofm)$ and
$\avgDelP(\sfm,\ofm)$. While the naive model produces an average absolute relative difference of approximately 10\% for
the first eigenvalue, the error jumps to 35\% 
\begin{figure}[H]
  \centerline{
    \includegraphics[width=0.5\textwidth]{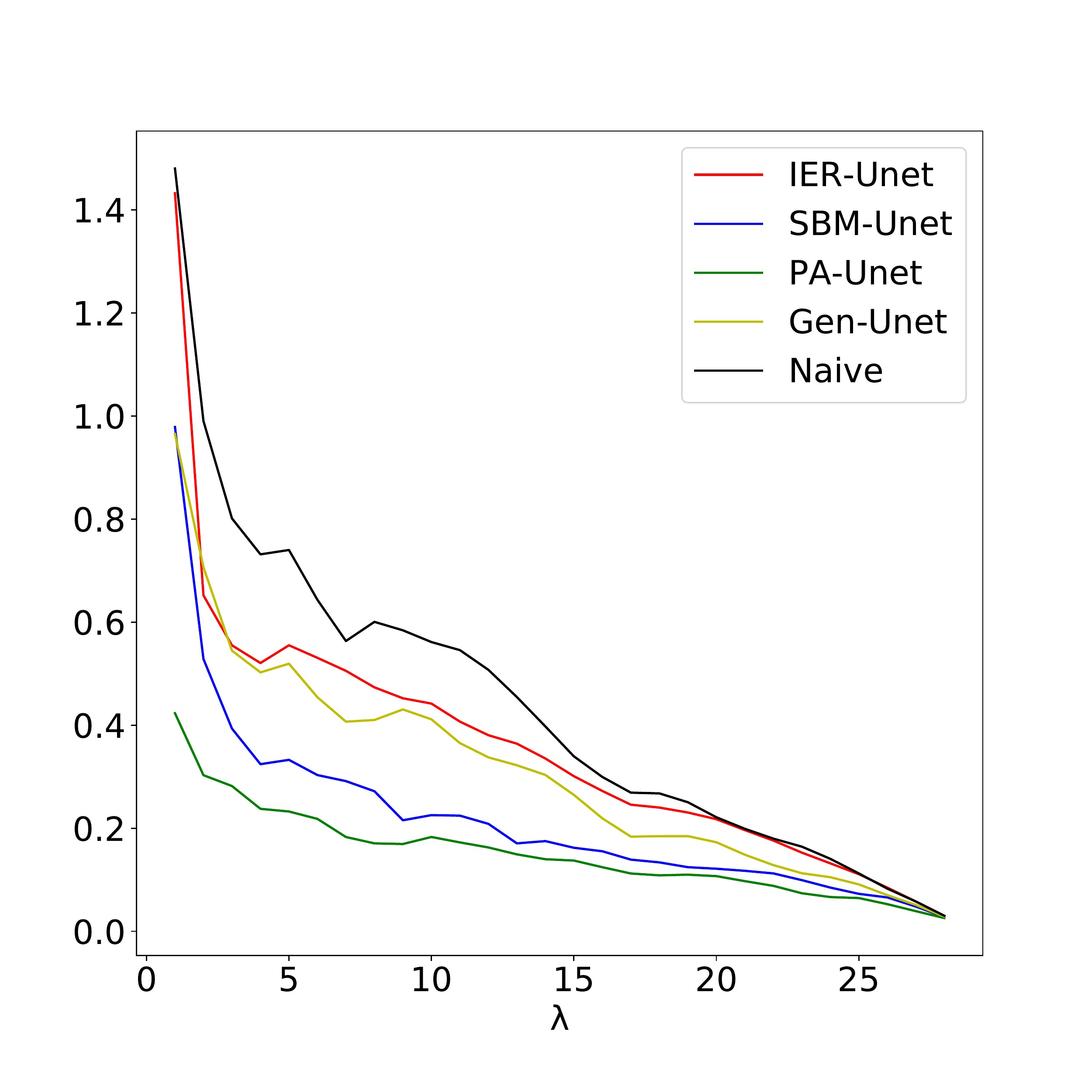}
    \hfill
    \includegraphics[width=0.5\textwidth]{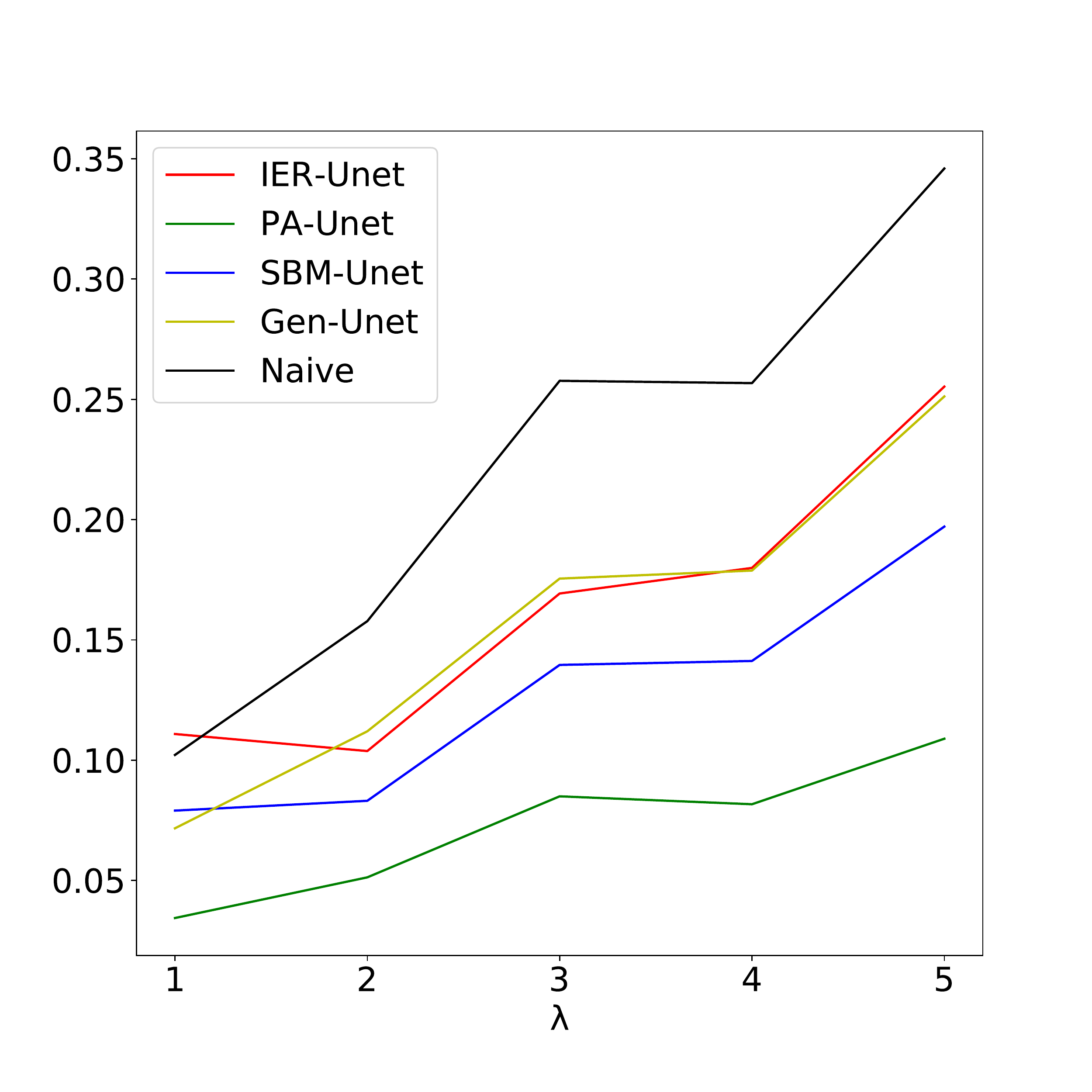}
  }
  \caption{Preferential attachment testing data: error between the eigenvalues of the true and predicted \fr mean for
    each model; absolute error $\avgDel(\sfm,\ofm)$ for the first 25 eigenvalues (left); relative error for the first
    five eigenvalues, $\avgDelP(\sfm,\ofm)$ (right).
    \label{fig9}}
\end{figure}

\begin{figure}[H]
  \centerline{
    \includegraphics[width=0.5\textwidth]{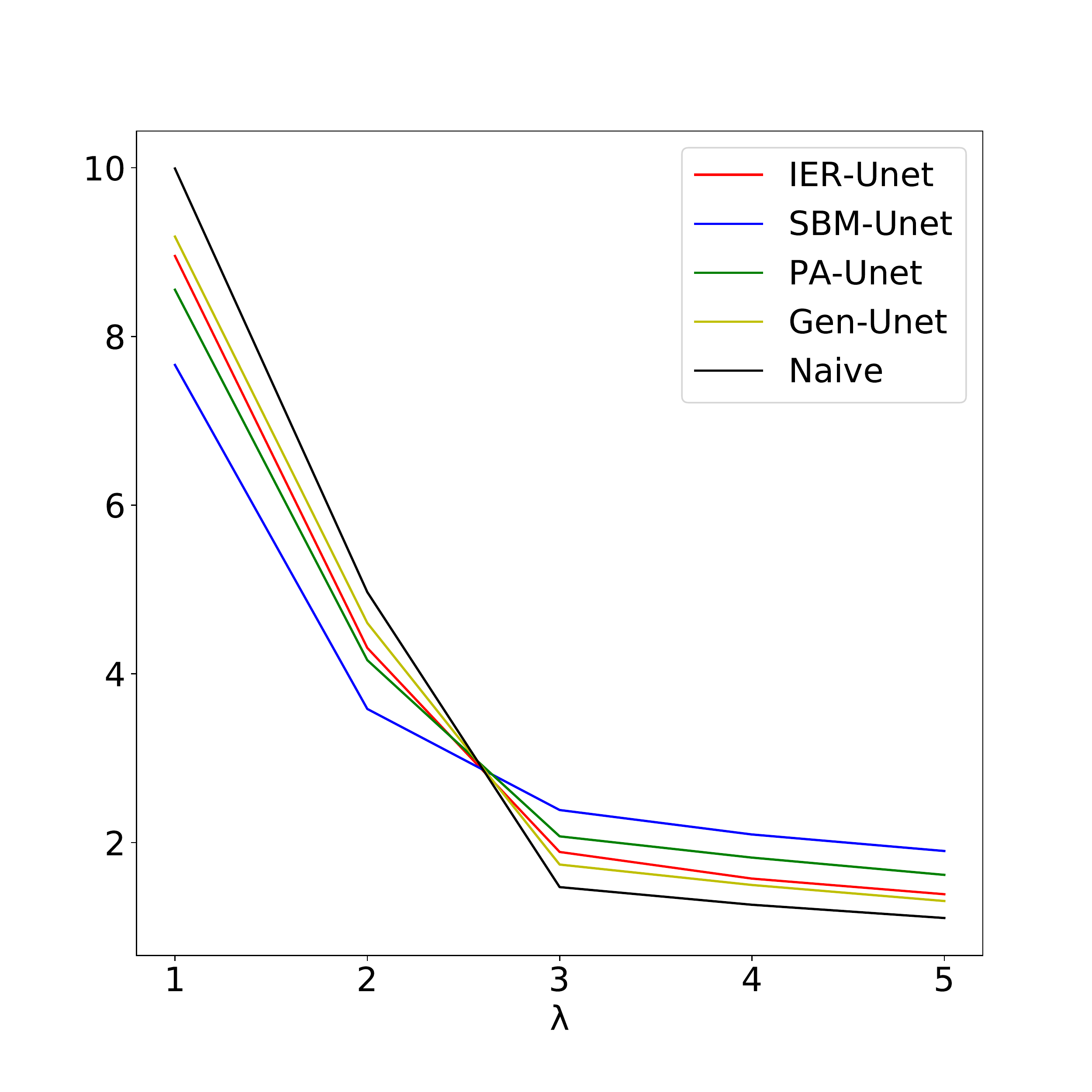}
    \hfill
    \includegraphics[width=0.5\textwidth]{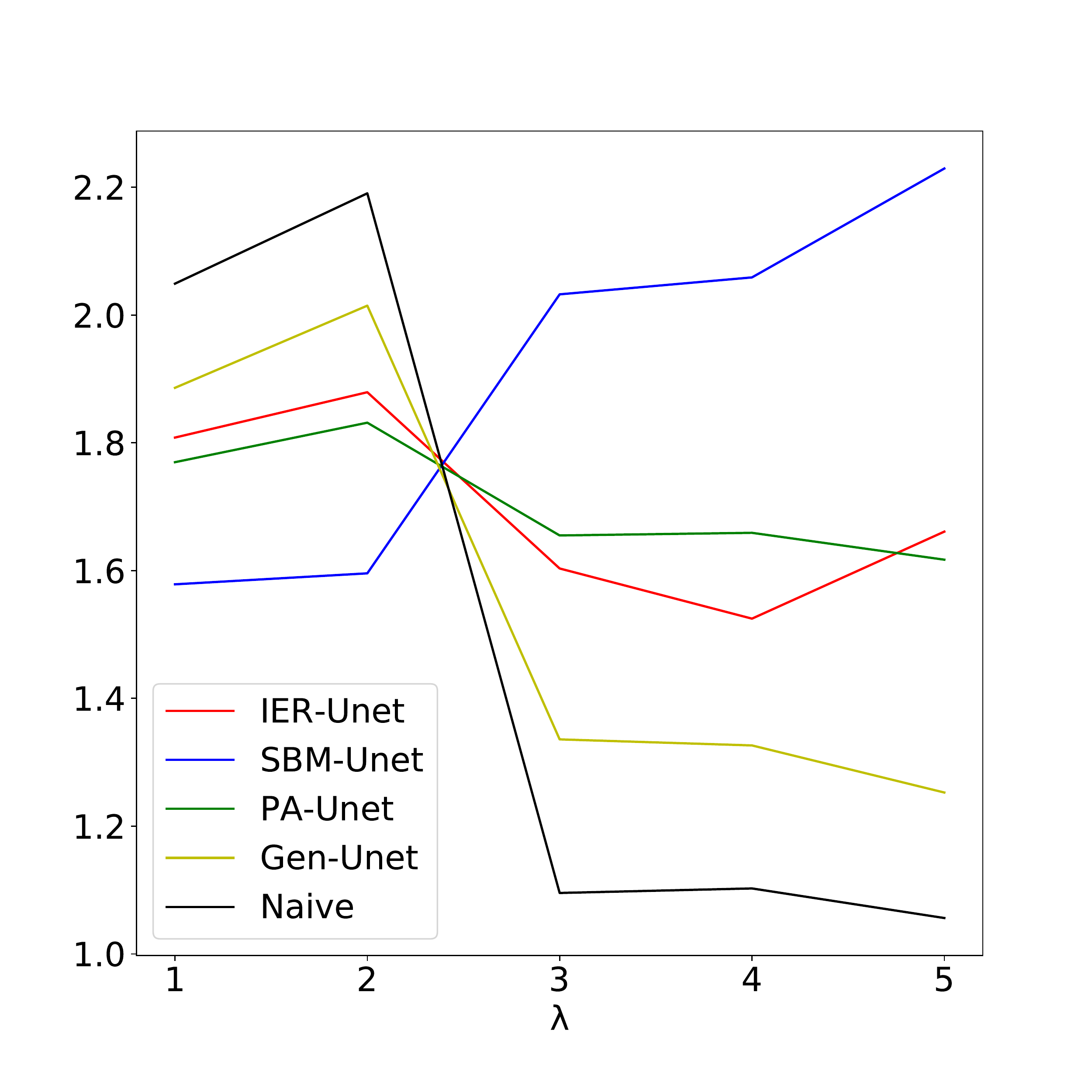}
  }
  \caption{Preferential attachment testing data: spectral difference between the eigenvalues of the predicted \fr mean
    and the sample mean eigenvalues; absolute difference $\avgDel(\blb(\ofm), \sE{\blb} )$ for the first 5 eigenvalues
    (left); relative difference for the first five eigenvalues, $\avgDelP(\blb(\ofm), \sE{\blb} )$ (right).
    \label{figA}}
\end{figure}
\noindent for the fifth eigenvalue (see Fig.~\ref{fig9}-right). In contrast, the error
created by the PA-Unet model is less than 3.5\% for the first eigenvalue, and is bounded by 10\% for the fifth
eigenvalue (see Fig~\ref{fig9}-right).

Our experiments demonstrate that all our other models perform remarkably well compared to the naive model model. This
results suggest that our algorithm can estimate the sample \fr mean for a wide variety of graph topology (not just IER,
or SBM). These results are promising as preferential attachment models aim to capture the "scale-free" property present
in many real world graph degree distributions.

The spectral difference between the eigenvalues of the predicted \fr mean and the sample mean eigenvalues are minimum
(across the first five eigenvalues) for the model trained with the SBM data, SBM-Unet, (see Fig.~\ref{figA}).  All of
our models result in smaller differences than the naive model, $\nfm$, until the third eigenvalue.  This result is
indeed very surprising considering that SBM-Unet was not trained on preferential attachment data so we would not expect
it to perform as well as it does. The SBM-Unet is probably able to reproduce the large hubs that are present in the
preferential attachment graphs, and which mimic the connectivity associated with the communities present in the SBM
training data. These large hubs correspond to the dominant eigenvalues of the adjacency matrix.

Turning our attention to the comparisons between degree distributions, we find that all our models produce on average 
values of the KL divergence between the true and predicted sample \fr mean graphs that are lower than those of the naive
model (see Fig.~\ref{tbl5}).

\begin{table}[H]
  \caption{KL divergence statistics for Preferential Attachment generated testing data}
  \label{tbl5}
  \begin{center}
    \begin{tabular}{lrr}
      \toprule 
      Model & Mean & Variance\\
      \midrule
      IER-Unet   & 0.4872 & 0.0496 \\
      SBM-Unet  & 0.3109 & 0.0250 \\
      PA-Unet   & 0.3598 & 0.0224 \\
      Gen-Unet  & 0.4208 & 0.0262 \\
      Naive  & 0.4970 & 0.0378 \\
      \botrule
    \end{tabular}
  \end{center}
\end{table}
\section{Conclusion}
We have described a fast method to compute the sample \fr mean graph using convolutional neural networks. The \fr mean
graph has become a fundamental concept for graph-valued machine learning. Our approach relies on a combination of two
key ideas: (i) stochastic block models with community of various sizes provide universal approximants to graphs, and
(ii) the computation of the sample \fr mean of stochastic block models (sampled from the same probability measure) can
be performed using simple averaging and nonlinear thresholding. We designed a convolutional network, which learns the
combined optimal approximation of the graphs in a sample along with the averaging and nonlinear thresholding that yields
the sample \fr mean. Our experiments on several ensembles of random graphs demonstrate that our method can reliably
predict the sample \fr mean.


\bmhead{Acknowledgments}
This work was supported by the National Science Foundation, CCF/CIF 1815971.

\begin{thebibliography}{55}
\providecommand{\natexlab}[1]{#1}
\providecommand{\url}[1]{{#1}}
\providecommand{\urlprefix}{URL }
\providecommand{\doi}[1]{\url{https://doi.org/#1}}
\providecommand{\eprint}[2][]{\url{#2}}
 \bibcommenthead

\bibitem[{Abbe(2018)}]{abbe18}
Abbe E (2018) Community detection and stochastic block models: Recent
  developments. Journal of Machine Learning Research 18(177):1--86

\bibitem[{Anderes et~al(2016)Anderes, Borgwardt, and Miller}]{anderes16}
Anderes E, Borgwardt S, Miller J (2016) Discrete \mbox{\mbox{wasserstein}}
  barycenters: Optimal transport for discrete data. Mathematical Methods of
  Operations Research 84(2):389--409

\bibitem[{Banks and Constantine(1998)}]{banks98}
Banks D, Constantine G (1998) Metric models for random graphs. Journal of
  Classification 15(2):199--223

\bibitem[{Barab{\'a}si and Albert(1999)}]{Barabasi1999}
Barab{\'a}si AL, Albert R (1999) Emergence of scaling in random networks.
  Science 286(5439):509--512. \doi{10.1.1.226.2025}

\bibitem[{Barbe et~al(2020)Barbe, Sebban, Gon{\c{c}}alves, Borgnat, and
  Gribonval}]{barbe20}
Barbe A, Sebban M, Gon{\c{c}}alves P, et~al (2020) Graph diffusion
  \mbox{Wasserstein} distances. In: Joint European Conference on Machine
  Learning and Knowledge Discovery in Databases, Springer, pp 577--592

\bibitem[{Bardaji et~al(2010{\natexlab{a}})Bardaji, Ferrer, and
  Sanfeliu}]{bardaji10b}
Bardaji I, Ferrer M, Sanfeliu A (2010{\natexlab{a}}) A comparison between two
  representatives of a set of graphs: median vs. barycenter graph. In: Joint
  IAPR International Workshops on Statistical Techniques in Pattern Recognition
  (SPR) and Structural and Syntactic Pattern Recognition (SSPR), Springer, pp
  149--158

\bibitem[{Bardaji et~al(2010{\natexlab{b}})Bardaji, Ferrer, and
  Sanfeliu}]{bardaji10a}
Bardaji I, Ferrer M, Sanfeliu A (2010{\natexlab{b}}) Computing the barycenter
  graph by means of the graph edit distance. In: 2010 20th International
  Conference on Pattern Recognition, IEEE, pp 962--965

\bibitem[{Bickel and Chen(2009)}]{bickel09}
Bickel PJ, Chen A (2009) A nonparametric view of network models and
  newman--girvan and other modularities. Proceedings of the National Academy of
  Sciences 106(50):21,068--21,073

\bibitem[{Bollobás et~al(2007)Bollobás, Janson, and Riordan}]{bollobas07}
Bollobás B, Janson S, Riordan O (2007) The phase transition in inhomogeneous
  random graphs. Random Structures \& Algorithms 31(1):3--122.
  \doi{https://doi.org/10.1002/rsa.20168}

\bibitem[{Boria et~al(2020)Boria, Negrevergne, and Yger}]{boria20}
Boria N, Negrevergne B, Yger F (2020) \mbox{Fr\'echet} mean computation in
  graph space through projected block gradient descent. In: ESANN 2020

\bibitem[{Brogat-Motte et~al(2022)Brogat-Motte, Flamary, Brouard, Rousu, and
  d’Alch{\'e} Buc}]{brogat22}
Brogat-Motte L, Flamary R, Brouard C, et~al (2022) Learning to predict graphs
  with fused \mbox{Gromov-Wasserstein} barycenters. In: International
  Conference on Machine Learning, PMLR, pp 2321--2335

\bibitem[{Cai et~al(2015)Cai, Ackerman, and Freer}]{cai15}
Cai D, Ackerman N, Freer C (2015) An iterative step-function estimator for
  graphons. arXiv preprint arXiv:14122129

\bibitem[{Chan and Airoldi(2014)}]{chan14}
Chan S, Airoldi E (2014) A consistent histogram estimator for exchangeable
  graph models. In: International Conference on Machine Learning, PMLR, pp
  208--216

\bibitem[{Chowdhury and M{\'e}moli(2019)}]{chowdhury19}
Chowdhury S, M{\'e}moli F (2019) The \mbox{Gromov-Wasserstein} distance between
  networks and stable network invariants. Information and Inference: A Journal
  of the IMA 8(4):757--787

\bibitem[{Donnat and Holmes(2018)}]{Donnat2018}
Donnat C, Holmes S (2018) Tracking network dynamics: A survey using graph
  distances. The Annals of Applied Statistics 12(2):971--1012

\bibitem[{Dubey and M{\"u}ller(2020)}]{dubey20}
Dubey P, M{\"u}ller HG (2020) \mbox{Fr\'echet} change-point detection. The
  Annals of Statistics 48(6):3312--3335

\bibitem[{Ferguson and Meyer(2022{\natexlab{a}})}]{ferguson22b}
Ferguson D, Meyer FG (2022{\natexlab{a}}) Computation of the sample
  \mbox{Fr\'echet} mean for sets of large graphs with applications to
  regression. In: Proceedings of the 2022 SIAM International Conference on Data
  Mining (SDM), SIAM, pp 379--387

\bibitem[{Ferguson and Meyer(2022{\natexlab{b}})}]{ferguson22}
Ferguson D, Meyer FG (2022{\natexlab{b}}) On the number of edges of the
  \mbox{Fr\'echet} mean and median graphs. In: Ribeiro P, Silva F, Mendes JF,
  et~al (eds) Network Science. Springer International Publishing, pp 26--40

\bibitem[{Ferguson and Meyer(2022{\natexlab{c}})}]{ferguson22a}
Ferguson D, Meyer FG (2022{\natexlab{c}}) Theoretical analysis and computation
  of the sample \mbox{Fr\'echet} mean for sets of large graphs based on
  spectral information. arXiv preprint arXiv:220105923

\bibitem[{Ferrer et~al(2010)Ferrer, Valveny, Serratosa, Riesen, and
  Bunke}]{ferrer10}
Ferrer M, Valveny E, Serratosa F, et~al (2010) Generalized median graph
  computation by means of graph embedding in vector spaces. Pattern Recognition
  43(4):1642--1655

\bibitem[{Fr\mbox{\'e}chet(1947)}]{frechet47}
Fr\mbox{\'e}chet M (1947) Les espaces abstraits et leur utilit{\'e} en
  statistique th{\'e}orique et m{\^e}me en statistique appliqu{\'e}e. Journal
  de la Soci{\'e}t{\'e} Fran{\c{c}}aise de Statistique 88:410--421

\bibitem[{Ginestet(2012)}]{ginestet12}
Ginestet CE (2012) Strong consistency of \mbox{Fr\'echet} sample mean sets for
  graph-valued random variables. arXiv preprint arXiv:12043183

\bibitem[{Ginestet et~al(2017)Ginestet, Li, Balachandran, Rosenberg, and
  Kolaczyk}]{ginestet17}
Ginestet CE, Li J, Balachandran P, et~al (2017) Hypothesis testing for network
  data in functional neuroimaging. The Annals of Applied Statistics
  11(2):725--750

\bibitem[{Gu et~al(2015)Gu, Hua, and Liu}]{gu15}
Gu J, Hua B, Liu S (2015) Spectral distances on graphs. Discrete Applied
  Mathematics 190:56--74

\bibitem[{Heinemann et~al(2022)Heinemann, Munk, and Zemel}]{heinemann22}
Heinemann F, Munk A, Zemel Y (2022) Randomized \mbox{\mbox{wasserstein}}
  barycenter computation: Resampling with statistical guarantees. SIAM Journal
  on Mathematics of Data Science 4(1):229--259

\bibitem[{Jain and Obermayer(2008)}]{jain08}
Jain B, Obermayer K (2008) On the sample mean of graphs. In: 2008 IEEE
  International Joint Conference on Neural Networks (IEEE World Congress on
  Computational Intelligence), IEEE, pp 993--1000

\bibitem[{Jain(2016{\natexlab{a}})}]{jain16a}
Jain BJ (2016{\natexlab{a}}) On the geometry of graph spaces. Discrete Applied
  Mathematics 214:126--144

\bibitem[{Jain(2016{\natexlab{b}})}]{jain16b}
Jain BJ (2016{\natexlab{b}}) Statistical graph space analysis. Pattern
  Recognition 60:802--812

\bibitem[{Jain and Obermayer(2009)}]{jain09}
Jain BJ, Obermayer K (2009) Algorithms for the sample mean of graphs. In:
  International Conference on Computer Analysis of Images and Patterns,
  Springer, pp 351--359

\bibitem[{Jain and Obermayer(2012)}]{jain12}
Jain BJ, Obermayer K (2012) Learning in \mbox{Riemannian} orbifolds.
  \eprint{1204.4294}

\bibitem[{Jiang et~al(2001)Jiang, Munger, and Bunke}]{jiang01}
Jiang X, Munger A, Bunke H (2001) On median graphs: properties, algorithms, and
  applications. IEEE Transactions on Pattern Analysis and Machine Intelligence
  23(10):1144--1151

\bibitem[{Josephs et~al(2021)Josephs, Li, and Kolaczyk}]{josephs21}
Josephs N, Li W, Kolaczyk ED (2021) Network recovery from unlabeled noisy
  samples. \eprint{2104.14952}

\bibitem[{Kingma and Ba(2015)}]{kingma15}
Kingma DP, Ba J (2015) Adam: A method for stochastic optimization. In: ICLR
  (Poster), \urlprefix\url{http://arxiv.org/abs/1412.6980}

\bibitem[{Kolaczyk et~al(2020)Kolaczyk, Lin, Rosenberg, Walters, Xu
  et~al}]{kolaczyk20}
Kolaczyk ED, Lin L, Rosenberg S, et~al (2020) Averages of unlabeled networks:
  Geometric characterization and asymptotic behavior. The Annals of Statistics
  48(1):514--538

\bibitem[{Kolouri et~al(2021)Kolouri, Naderializadeh, Rohde, and
  Hoffmann}]{kolouri21}
Kolouri S, Naderializadeh N, Rohde GK, et~al (2021) \mbox{\mbox{wasserstein}}
  embedding for graph learning. In: International Conference on Learning
  Representations

\bibitem[{Kullback and Leibler(1951)}]{kullback51}
Kullback S, Leibler RA (1951) On information and sufficiency. The annals of
  mathematical statistics 22(1):79--86

\bibitem[{Leskovec et~al(2009)Leskovec, Lang, Dasgupta, and
  Mahoney}]{leskovec09}
Leskovec J, Lang KJ, Dasgupta A, et~al (2009) Community structure in large
  networks: Natural cluster sizes and the absence of large well-defined
  clusters. Internet Mathematics 6(1):29--123

\bibitem[{Lunag{\'o}mez et~al(2020)Lunag{\'o}mez, Olhede, and
  Wolfe}]{lunagomez20}
Lunag{\'o}mez S, Olhede SC, Wolfe PJ (2020) Modeling network populations via
  graph distances. Journal of the American Statistical Association pp 1--18

\bibitem[{McKay and Piperno(2014)}]{mckay2014}
McKay B, Piperno A (2014) Practical graph isomorphism, ii. Journal of Symbolic
  Computation 60:94--112

\bibitem[{M{\'e}tivier et~al(2019)M{\'e}tivier, Brossier, Merigot, and
  Oudet}]{metivier19}
M{\'e}tivier L, Brossier R, Merigot Q, et~al (2019) A graph space optimal
  transport distance as a generalization of lp distances: application to a
  seismic imaging inverse problem. Inverse Problems 35(8):085,001

\bibitem[{Meyer(2021)}]{meyer22b}
Meyer FG (2021) The \mbox{Fr\'echet} mean of inhomogeneous random graphs. In:
  International Conference on Complex Networks and Their Applications.
  Springer, pp 207--219

\bibitem[{Meyer(2022)}]{meyer22}
Meyer FG (2022) The \mbox{Fr\'echet} mean of inhomogeneous random graphs. In:
  Complex Networks {\&} Their Applications X. Springer, pp 207--219

\bibitem[{Morris et~al(2020)Morris, Kriege, Bause, Kersting, Mutzel, and
  Neumann}]{morris20}
Morris C, Kriege NM, Bause F, et~al (2020) Tudataset: A collection of benchmark
  datasets for learning with graphs. In: ICML 2020 Workshop on Graph
  Representation Learning and Beyond (GRL+ 2020),
  \urlprefix\url{www.graphlearning.io}, \eprint{2007.08663}

\bibitem[{Olhede and Wolfe(2014)}]{olhede14}
Olhede SC, Wolfe PJ (2014) Network histograms and universality of blockmodel
  approximation. Proceedings of the National Academy of Sciences
  111(41):14,722--14,727

\bibitem[{Patterson(2021)}]{patterson21}
Patterson E (2021) Hausdorff and \mbox{\mbox{wasserstein}} metrics on graphs
  and other structured data. Information and Inference: A Journal of the IMA
  10(4):1209--1249

\bibitem[{Preuer et~al(2018)Preuer, Renz, Unterthiner, Hochreiter, and
  Klambauer}]{preuer18}
Preuer K, Renz P, Unterthiner T, et~al (2018) \mbox{\mbox{fr\'echet} chemnet}
  distance: a metric for generative models for molecules in drug discovery.
  Journal of chemical information and modeling 58(9):1736--1741

\bibitem[{Ronneberger et~al(2015)Ronneberger, Fischer, and
  Brox}]{ronneberger15}
Ronneberger O, Fischer P, Brox T (2015) U-net: Convolutional networks for
  biomedical image segmentation. In: International Conference on Medical image
  computing and computer-assisted intervention, Springer, pp 234--241

\bibitem[{Sanchez(2022)}]{frechet-adam-sanchez22}
Sanchez A (2022) Estimation of the sample \mbox{Fr\'echet} mean: A
  convolutional neural network approach.
  \url{https://github.com/mra717/Estimation_of_Sample_Fr_Mean}

\bibitem[{Simou et~al(2020)Simou, Thanou, and Frossard}]{simou20}
Simou E, Thanou D, Frossard P (2020) Node2coords: Graph representation learning
  with \mbox{Wasserstein} barycenters. IEEE Transactions on Signal and
  Information Processing over Networks 7:17--29

\bibitem[{Solomon et~al(2015)Solomon, De~Goes, Peyr{\'e}, Cuturi, Butscher,
  Nguyen, Du, and Guibas}]{solomon15}
Solomon J, De~Goes F, Peyr{\'e} G, et~al (2015) Convolutional
  \mbox{Wasserstein} distances: Efficient optimal transportation on geometric
  domains. ACM Transactions on Graphics (ToG) 34(4):1--11

\bibitem[{Vayer et~al(2020)Vayer, Chapel, Flamary, Tavenard, and
  Courty}]{vayer20}
Vayer T, Chapel L, Flamary R, et~al (2020) Fused \mbox{Gromov-Wasserstein}
  distance for structured objects. Algorithms 13(9):212

\bibitem[{Wei(2021)}]{wei21}
Wei S (2021) Multidimensional graph trend filtering. PhD thesis, University of
  California, Davis

\bibitem[{Wei et~al(2018)Wei, Madrid-Padilla, and Sharpnack}]{wei18}
Wei S, Madrid-Padilla OH, Sharpnack J (2018) Distributed cartesian power graph
  segmentation for graphon estimation. arXiv preprint arXiv:180509978

\bibitem[{Wills and Meyer(2020)}]{wills20c}
Wills P, Meyer FG (2020) Metrics for graph comparison: A practitioner’s
  guide. PLOS ONE 15(2):1--54. \doi{10.1371/journal.pone.0228728},
  \urlprefix\url{https://doi.org/10.1371/journal.pone.0228728}

\bibitem[{Wolfe and Olhede(2013)}]{wolfe13}
Wolfe PJ, Olhede SC (2013) Nonparametric graphon estimation.
  \doi{10.48550/ARXIV.1309.5936},
  \urlprefix\url{https://arxiv.org/abs/1309.5936}

\end{thebibliography}

\end{document}